\documentclass[conference]{IEEEtran}
\usepackage{times}

\usepackage{wrapfig}
\usepackage[numbers]{natbib}
\usepackage{multicol}

\usepackage{dblfloatfix}
\usepackage{makecell}
\usepackage{longtable}
\newcommand{\ourmodel}{\textsc{CREStE}}

\newcommand{\seclabel}[1]{\label{sec:#1}}
\newcommand{\secref}[1]{Sec.~\ref{sec:#1}}
\newcommand{\sseclabel}[1]{\label{ssec:#1}}
\newcommand{\ssecref}[1]{Sec.~\ref{ssec:#1}}

\newcommand{\figlabel}[1]{\label{fig:#1}}
\newcommand{\figref}[1]{Fig.~\ref{fig:#1}}
\newcommand{\tablabel}[1]{\label{tab:#1}}
\newcommand{\tabref}[1]{Table~\ref{tab:#1}}
\newcommand{\eqnlabel}[1]{\label{eq:#1}}
\newcommand{\eqnref}[1]{Eq.~\ref{eq:#1}}

\usepackage{multirow}
\usepackage[table,xcdraw]{xcolor} 
\usepackage{array}               
\usepackage{booktabs}            
\usepackage{amsmath}  
\usepackage{amsfonts} 
\usepackage{amssymb}  
\usepackage{graphicx}
\usepackage{caption}
\usepackage{comment}
\usepackage{hyperref}

\newcommand{\jb}[1]{[{\textbf{\textcolor{red}{Joydeep: #1}}}]}

\newcommand{\harshit}[1]{\textcolor{orange}{\small{\bf [HS: #1]}}}
\newcommand{\akz}[1]{[{\textbf{\textcolor{magenta}{Arthur: #1}}}]}
\newcommand{\mc}[1]{\multicolumn{1}{l|}{#1}}

\begin{document}

\title{\ourmodel{}: Scalable Mapless Navigation with Internet Scale Priors and Counterfactual Guidance}

\author{
    Arthur Zhang, Harshit Sikchi, Amy Zhang, Joydeep Biswas \\
    \small The University of Texas at Austin
}


\twocolumn[{%
\renewcommand\twocolumn[1][]{#1}%
\maketitle
\begin{center}
    \vspace{-20pt}
    \centering
	\includegraphics[width=1.0\textwidth]{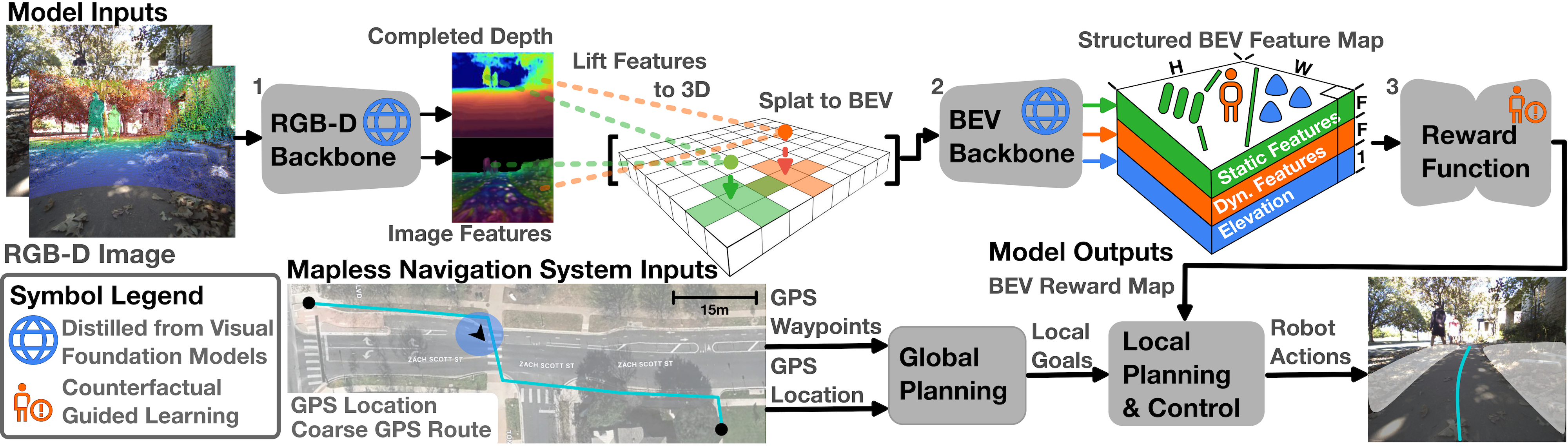}
	\captionof{figure}{\ourmodel{} learns a perceptual encoder$^{1,2}$ and reward function$^3$ to predict structured bird's eye view (BEV) feature and reward maps for navigation. Our model inherits generalization and robustness from visual foundation models and learns expert-aligned rewards using our counterfactual-guided learning framework. We integrate \ourmodel{} in a modular navigation system that uses coarse GPS guidance and rewards to reach navigation goals safely.}
    \label{fig:header}
    \figlabel{architectureoverview}
\end{center}%
}]

\IEEEpeerreviewmaketitle

\begin{abstract}
We introduce \ourmodel{}, a scalable learning-based mapless navigation framework to address the open-world generalization and robustness challenges of outdoor urban navigation.
Key to achieving this is learning perceptual representations that generalize to open-set factors (e.g. novel semantic classes, terrains, dynamic entities) and inferring expert-aligned navigation costs from limited demonstrations. 
\ourmodel{} addresses both these issues, introducing 1) a visual foundation model (VFM) distillation objective for learning open-set 
structured bird’s-eye-view perceptual representations,  and 2) counterfactual inverse reinforcement learning (IRL), a novel active learning formulation that uses counterfactual trajectory demonstrations to reason about the most important cues when inferring navigation costs. 
We evaluate \ourmodel{} on the task of kilometer-scale mapless navigation in a variety of city, offroad, and residential environments and find that it outperforms all state-of-the-art approaches with 70\% fewer human interventions, including a 2-kilometer mission in an unseen environment with just 1 intervention; showcasing its robustness and effectiveness for long-horizon mapless navigation. Videos and additional materials can be found on the project page: \noindent\textbf{\texttt{\href{https://amrl.cs.utexas.edu/creste}{https://amrl.cs.utexas.edu/creste}}}.

\end{abstract}


\section{Introduction}
\seclabel{introduction}

Mapless navigation is the task of reaching user-specified goals without high-definition (HD) maps and precise navigation waypoints. Mapless approaches plan routes using egocentric sensor observations (e.g. RGB images, point clouds, GPS), coarse waypoints from public routing services, and satellite imagery. These solutions demonstrate promise as a scalable alternative to conventional map-centric approaches: enabling generalization to unforeseen factors (e.g. curb ramps and foliage) and dynamic entities (e.g. pedestrians and strollers) while reducing map maintenance overhead and reliance on pre-defined routes.

Traditional geometric-only mapless solutions~\cite{biswas2016thousand, thrun1999minerva} exhibit robust generalization when it is only necessary to consider costs for geometric factors like static obstacles. However, open-world navigation demands perception systems that perceive an open set of factors unknown apriori, ranging from terrain preferences (grass vs. concrete) to semantic cues (crosswalks and crossing signs). Furthermore, it requires the ability to identify the most salient features in the scene,  and how they influence the choice of paths.

Learning-based approaches are a scalable alternative that considers factors beyond geometry, but must overcome data scarcity, robustness, and generalization challenges to achieve similar reliability. These approaches broadly consist of single-factor perception, hand-curated multi-factor perception, end-to-end learning, and zero-shot pre-trained large language model (LLM)/visual language model (VLM) transfer.  While single-factor~\cite{mao2024pacer, weerakoon2022terp, thrun1999minerva} and multi-factor~\cite{meng2023terrainnet, roth2024viplanner} methods learn representations that consider relevant navigation factors (e.g. geometry, terrain, semantics), they rely on a hand-curated list of semantic classes and terrains that limit generalization to unseen classes. End-to-end methods~\cite{shah2022viking, shah2023vint, shah2021ving} alleviate this by jointly learning the representation and policy from expert demonstrations, but are prone to overfitting without large-scale robot datasets. While recent works~\cite{weerakoon2024behav, nasiriany2024pivot} demonstrate that pre-trained LLMs and VLMs can reason about expert-aligned behavior without large-scale robot datasets, we empirically demonstrate that they are poorly attuned to urban navigation, leading to brittle zero-shot transfer in complex scenes. 

We address the aforementioned limitations with \ourmodel{}, \textbf{Counterfactuals for Reward Enhancement with Structured Embeddings}, a novel approach that learns open-set representations and navigation costs/rewards for mapless urban navigation. To learn open-set representations, \ourmodel{} combines prior work on bird's eye view (BEV) representation learning~\cite{meng2023terrainnet} and visual foundation models (VFMs)~\cite{oquab2023dinov2, ravi2024sam} to learn BEV map representations with inherited real-world robustness and open-set semantic knowledge. We unify priors from multiple foundation models by distilling image features from Dinov2~\cite{oquab2023dinov2} and refining these features in BEV using SAM2~\cite{ravi2024sam} instance labels. \ourmodel{} infers expert-aligned navigation costs without large-scale demonstrations by leveraging counterfactuals to guide learning, where counterfactuals hold all other variables constant except for the path taken from the start to the end goal. Unlike conventional preference learning, our proposed counterfactual inverse reinforcement learning (IRL) objective explicitly minimizes rewards along paths that exhibit undesirable behavior (e.g. veering off crosswalks, driving off curb ramps, etc.), enabling operators to correct robot behavior by providing offline counterfactual feedback. We summarize the main contributions of our work as follows:

\begin{itemize}
    \item \textbf{Representation Learning Through Model Distillation.} A new model architecture and distillation objective for distilling navigation priors from visual foundation models to a lightweight image to BEV map backbone.  
    \item \textbf{Counterfactually Aligned Rewards.} An active learning framework and counterfactual IRL formulation for reward alignment using counterfactual and expert demonstrations.
\end{itemize}

We demonstrate our approach's effectiveness through real-world kilometer-scale navigation experiments in urban environments with off-road terrain, elevated walkways, sidewalks, intersections, and other challenging conditions. In total, we evaluate in \textbf{6} distinct seen and unseen geographic areas and significantly outperform existing state-of-the-art imitation learning, inverse reinforcement learning, and heuristic-based methods on the task of mapless navigation.

\section{Related Work}

In this section, we position our work within the broader context of methods that perform mapless navigation by learning representations and policies for safe local path planning. Underlying these methods are two key challenges: learning robust perceptual representations that encode an open set of navigation factors, and learning planning modules that can identify and reason about the most important factors to plan safe paths.

Single~\cite{karnan2023sterling, mao2024pacer, sikand2022visual} and multi-factor~\cite{gosala2023skyeye, meng2023terrainnet, roth2024viplanner} approaches learn perceptual representations grounded in the planning horizon with elevation, terrain, and semantic factors, and rely on expert tuning to balance the relative costs of each factor. While this enables joint reasoning about multiple factors, it assumes the full set of semantic classes and terrains are known apriori, generalizing poorly to unexpected semantic classes, terrains, or other factors not seen during training. Furthermore, they require expert tuning of complex multi-objective cost functions for each environment.

End-to-end methods jointly learn a representation and policy using either behavior cloning (BC), RL, or IRL. BC methods~\cite{kendall2019learning, shah2023vint} learn task-specific representations for reasoning about how to mimic the expert policy. These approaches scale effectively given large-scale datasets with expert demonstrations, but are prone to overfitting otherwise. RL~\cite{hirose2024lelan} and IRL~\cite{wulfmeier2016watch} methods also mimic the expert policy, but design handcrafted reward functions (e.g. minimize vibrations~\cite{kahn2021badgr} and likelihood of interventions~\cite{kahn2021land}) or preference rankings~\cite{sikchi2022ranking, biyik2022learning} to guide representation and policy learning. These methods efficiently learn expert-aligned behaviors at the cost of careful reward tuning or preference enumeration. 


An emerging class of methods~\cite{nasiriany2024pivot, weerakoon2024behav} leverages pre-trained factors present in VLMs and LLMs, large foundation models pre-trained on internet-scale data, zero-shot for navigation. These foundation models leverage geographic hints in the form of satellite imagery and topological maps, along with image observations and text instructions to predict safe local waypoints for analytical planning and control methods. While promising, it is not well understood how to best steer VLMs/LLMs to reason about the most important navigation factors for navigation tasks.

The aforementioned works either rely on large-scale robot datasets for open-set generalization or complex reward design to learn expert-aligned behavior. However, constructing large robot datasets and expressive multi-term objectives is prohibitively difficult for open-world settings. In contrast, \ourmodel{} learns open-set representations by directly distilling knowledge from VFMs, and expert-aligned reward functions by using counterfactuals as a unified language for conveying navigation constraints. Unlike counterfactual-based methods like VRL-PAP~\cite{sikand2022visual}, our counterfactuals specify arbitrary navigation constraints beyond terrains and can be readily obtained offline, enabling a data flywheel for continuous improvement with additional deployments. Importantly, our reward learning approach is distinct from preference-learning methods, as counterfactuals are a more specific form of feedback than preferences, which can be between any two trajectories.

\section{The Mapless Urban Navigation Problem}
\seclabel{background}

We now develop the mapless navigation problem for urban environments. We first formulate the path planning problem in this context in~\ssecref{background:pathplanning} and then discuss the problem of learning general expert-aligned costs in~\ssecref{background:preferencecosts}. Finally,~\ssecref{background:challenges} highlights key challenges in our problem formulation that this work addresses. In this work, we refer to costs as negated rewards and use them interchangeably.

\subsection{Path Planning for Mapless Navigation}
\sseclabel{background:pathplanning}

For each timestep, we assume the robot has access to the current observation $o_t$ and pose $x_t \in \mathcal{X}$ in the global frame, where the robot state space $\mathcal{X}$ lies in $SE(2)$ for ground vehicles. The robot must plan a finite trajectory $\Gamma_S = [x_t, ..., x_{t:t+S}]$ from $x_t$ to goal $G$, either given by the user or obtained from public routing services. $\Gamma_S$ consists of S current and future states $x\in \mathcal{X}$ which minimize the following objective function:
\begin{equation}
\Gamma_S = \text{arg}_\Gamma \text{min} || x_{t+S} - G || + \lambda \mathcal{J}(\Gamma),
\end{equation}
where $||x_{t+S} - G||$ is the distance between the final state $x_{t+S}$ and $G$, and $\mathcal{J}(\Gamma)$ is a cost function for path planning scaled by a relative weight $\lambda$. In mapless urban environments, the robot pose $x_t$ and goal $G$ may be highly noisy, causing $\mathcal{J}(\Gamma)$'s importance to vary dynamically across time. 

\subsection{General Preference-Aligned Cost Functions}
\sseclabel{background:preferencecosts}

To properly define our cost function $\mathcal{J}(\Gamma)$, we first need to introduce the notion of an observation function $\Theta : \mathcal{O} \rightarrow \mathcal{T}$ that maps observations $o_t$ to a joint embedding space $\mathcal{T}$ that encapsulates a set of relevant factors for navigation in the world. In general, the sufficient set of factors is unknown but can be approximated for each environment. Let $T : \mathcal{X} \rightarrow \mathcal{T}$ be a function that maps a robot pose $x\in\mathcal{X}$ to a feature $\tau\in\mathcal{T}$, where $\tau$ captures the relevant set of factors necessary to reason about $\mathcal{J}(\Gamma)$. The relationship and set of relevant factors may consist of geometric, semantic, and social costs as follows:
\begin{equation}
\begin{aligned}
\mathcal{J}(\Gamma) &= \sigma( \mathcal{J}_\text{geometric}(\Gamma), \mathcal{J}_\text{semantic}(\Gamma), \mathcal{J}_\text{social}(\Gamma))
\end{aligned}
\end{equation}
where $\sigma$ is a function that combines individual cost terms.

We assume the operator has an underlying true cost function $H: \mathcal{T} \rightarrow \mathbb{R}^{0+}$ mapping the sufficient set of factors to scalar real-valued costs based on their preferences. Let $H \in \mathcal{H}$, where $\mathcal{H}$ is the continuous space of underlying cost functions. In general, $H$ is unknown and often depends on the robot embodiment, environment, and task. For mapless navigation, we define this task to be goal-reaching.

\subsection{Open Challenges}
\sseclabel{background:challenges}

In this work, we are concerned with learning both $\Theta(o_t)$ and $\mathcal{J}(\Gamma)$. This is difficult as the sufficient set of factors for navigating diverse, urban environments is unknown and the relationship between factors may be highly nonlinear. Our approach to learning $\Theta(o_t)$ distills features from VFMs, which provide a breadth of factors including but not limited to: geometry, semantics, and entities. This promotes learning a joint distribution $\mathcal{T}$ sufficient for mapless urban navigation. Furthermore, we propose a counterfactual-based framework for learning $\mathcal{J}(\Gamma)$, which becomes important when dealing with complex feature distributions $\mathcal{T}$ and nonlinear relationships between factors $\sigma$.

\section{Approach}
\seclabel{approach}

\begin{figure*}[!ht]
    \centering
    \includegraphics[width=\linewidth]{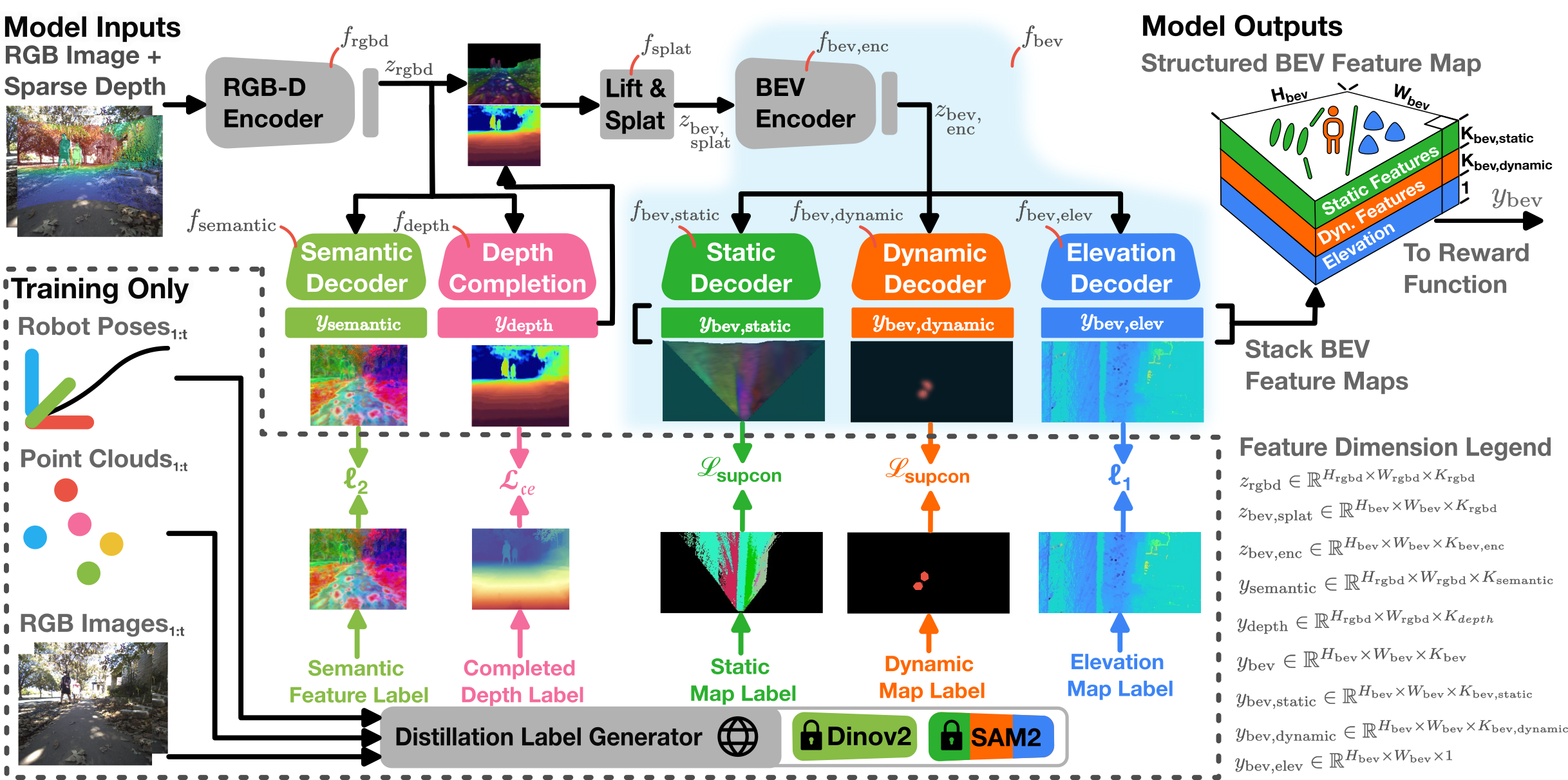}
    \caption{Model architecture and training procedure for the \ourmodel{} perceptual encoder $\Theta$. Using a RGB and sparse depth image, $f_\text{rgbd}$ extracts image features $z_\text{rgbd}$ and performs depth completion. Next, we lift and splat $z_\text{rgbd}$ to an unstructured BEV feature map before predicting continuous static panoptic, dynamic panoptic, and elevation features. Finally, we stack the predicted features to construct a structured BEV feature map for our learned reward function. We supervise $\Theta$ using semantic feature maps, completed depth, and BEV map labels generated by our SegmentAnythingv2~\cite{kirillov2023segany} and Dinov2~\cite{oquab2023dinov2}-powered distillation label generator. We define the dimensions for important feature maps in the Feature Dimension Legend on the bottom right.}
    \figlabel{perceptionpipeline}
    \vspace{-10pt}
\end{figure*}

\ourmodel{} is a modular approach with two key components that can be trained end-to-end: 1) A perceptual encoder $\Theta(o_\text{rgb, t}$, $o_\text{depth, t})$ that takes the robot's current RGB and sparse depth observation and predicts a completed depth image $y_\text{depth, t}$ and structured BEV feature map $y_\text{bev, t}$; 2) A reward function $r_\phi(y_\text{bev, t})$ that takes $y_\text{bev, t}$ and outputs a BEV scalar reward map $y_\text{reward, t}$. In the remainder of this section, we describe the following: 1) \ssecref{crestemodelarchitecture} - The \ourmodel{} model architecture and 2) \secref{crestetrainingprocedure} - The \ourmodel{} training procedure, where \ssecref{perceptiontrainingprocedure} presents our VFM distillation objective for $\Theta$ and \ssecref{rewardfunctionlearning} presents counterfactual IRL and our active reward learning framework for learning $r_\phi$.

\subsection{\ourmodel{} Model Architecture}
\sseclabel{crestemodelarchitecture}

\subsubsection{Perceptual Encoder Model Architecture}
\sseclabel{perceptionmodelarchitecture}

Our 25.5M parameter perceptual encoder $\Theta$ draws inspiration from the TerrainNet~\cite{meng2023terrainnet} backbone, which trains a RGB-D encoder $f_\text{rgbd}$ to predict a latent feature map $z_\text{rgbd}$ using an EfficientNet-B0~\cite{tan2019efficientnet} encoder and a completed depth map $y_\text{depth}$ using a depth completion head $f_\text{depth}(z_\text{rgbd})$. Like TerrainNet, we train a lift-splat module $f_\text{splat}(z_\text{rgbd}, y_\text{depth})$ to lift latent features to 3D and ``splat" them to an unstructured BEV feature map $z_\text{bev, splat}$. Finally, we pass $z_\text{bev, splat}$ to a BEV inpainting backbone $f_\text{bev}$ that uses a shared U-Net~\cite{ronneberger2015u} encoder and separate decoders to predict a structured feature map consisting of separate semantic and elevation layers.

Building on TerrainNet, we make two key architectural modifications. Our first modification, semantic decoder $f_\text{semantic}$, promotes learning semantic and geometric-aware features by regressing image features from Dinov2~\cite{oquab2023dinov2} using latent image features $z_\text{rgbd}$. This design is analogous to model distillation~\cite{ranzinger2024radio} and allows our RGB-D encoder $f_\text{rgbd}$ to inherit properties from VFMs like robustness to perceptual aliasing and open-set semantic understanding. 

Our second modification stems from the observation that Dinov2 features alone lack the entity understanding needed to backproject and inpaint features in heavily occluded urban scenes with noisy depth predictions. Thus, we supplement $f_\text{bev}$ with two panoptic map decoders $f_\text{bev,static}$ and $f_\text{bev,dynamic}$ to ensure that predicted BEV feature maps $y_\text{bev}$ are consistent with BEV instance maps from SegmentAnythingv2 (SAM2)~\cite{ravi2024sam}. We use Supervised Contrastive Loss~\cite{khosla2020supervised} to optimize $y_\text{bev}$ such that features belonging to the same instance are closer in embedding space than features belonging to different instances. Combined with $f_\text{semantic}$, our modifications synergistically unify the strengths of Dinov2 and SAM2 to learn open-set semantic, geometric, and instance-aware representations grounded in the local planning horizon. Altogether, our structured BEV feature map $y_\text{bev}$ consists of three layers stacked along the channel dimension: 1) Static panoptic feature map $y_\text{bev, static}$, 2) Dynamic panoptic feature map $y_\text{bev, dynamic}$, and 3) Elevation map $y_\text{bev, elev}$.

\subsubsection{Reward Function Model Architecture}
\sseclabel{rewardmodelarchitecture}

To ensure our reward function enforces spatial invariance and considers multi-scale features, we implement $r_\phi$ using a 0.5M parameter Multi-Scale Fully Convolutional Network (MS FCN) first used by Wulfmeier et al.~\cite{wulfmeier2016watch}. We supervise $r_\phi$ using our counterfactual IRL loss, which we describe along with our training procedure for $\Theta$ in the next section.

\subsection{\ourmodel{} Training Procedure}
\seclabel{crestetrainingprocedure}

We train \ourmodel{} by first optimizing $\Theta$ and freezing the parameters before training $r_\phi$ using our counterfactual-based active reward learning framework. Altogether, our full learning objective is:
\begin{equation}
\mathcal{L}_{\ourmodel{}} = \mathcal{L}_\text{rgbd} + \mathcal{L}_\text{bev} + \mathcal{L}_\text{IRL}
\end{equation}
where $\mathcal{L}_\text{rgbd}$ and $\mathcal{L}_\text{bev}$ supervise $\Theta$ and $\mathcal{L}_\text{IRL}$ supervise $r_\phi$. \ssecref{perceptiontrainingprocedure} breaks down the our training and label generation procedure for supervising $\Theta$ and \ssecref{rewardfunctionlearning} defines our counterfactual IRL objective $\mathcal{L}_\text{IRL}$ and active reward learning framework for training $r_\phi$.

\subsubsection{Training the Perceptual Encoder $\Theta$}
\sseclabel{perceptiontrainingprocedure}

We jointly train $f_\text{rgbd}$ via backpropagation from $f_\text{semantic}$ and $f_\text{depth}$. The semantic decoder head $f_\text{semantic}$ is supervised via an MSE loss that minimizes the error between the predicted and ground truth feature maps $y_\text{semantic}$ and $\hat{y}_\text{semantic}$. The depth completion head $f_\text{depth}$ is supervised using a cross-entropy classification loss $\mathcal{L}_{\mathcal{C}\mathcal{E}}$, where the ground truth depth $\hat{y}_\text{depth}$ is uniformly discretized into bins. Empirically, this captures depth discontinuities better than regression~\cite{reading2021categorical}. Altogether, the training objective for the RGB-D backbone is:
\begin{equation}
\begin{aligned}
\mathcal{L}_\text{rgbd} &= \alpha_1 l_2(y_\text{semantic}, \hat{y}_\text{semantic}) + \alpha_2 \mathcal{L}_{\mathcal{C}\mathcal{E}} (y_\text{depth}, \hat{y}_\text{depth})
\end{aligned}
\end{equation}
where $\alpha_1$ and $\alpha_2$ are tunable hyperparameters.

We supervise $f_\text{bev}$ with three losses, one for each BEV decoder head, using ground truth static instance maps $\hat{y}_\text{bev, static}$, dynamic instance maps $\hat{y}_\text{bev, dynamic}$, and elevation maps $\hat{y}_\text{bev, elev}$. We use Supervised Contrastive Loss~\cite{khosla2020supervised} ($\mathcal{L}_\text{contrastive}$) for training $f_\text{bev, static}$ and $f_\text{bev, dynamic}$ to predict continuous feature maps from discrete instance labels $\hat{y}_\text{bev, static}$ and $\hat{y}_\text{bev, dynamic}$. We train  $f_\text{bev, elev}$ to predict $\hat{y}_\text{bev, elev}$ using $l_1$ regression loss. Our full training objective for $f_\text{bev}$ is:
\begin{equation}
\begin{aligned}
\mathcal{L}_\text{bev} &= \beta_1 \mathcal{L}_{\mathrm{contrastive}}(y_\text{static}, \hat{y}_\text{static}) + \\
    &\beta_2 \mathcal{L}_{\mathrm{contrastive}}(y_\text{dynamic}, \hat{y}_\text{dynamic})+
    \beta_3 l_1(y_\text{elev}, \hat{y}_\text{elev})
\end{aligned}
\end{equation}
where $\beta_1$, $\beta_2$, and $\beta_3$ are tunable hyperparameters. We refer readers to Appendix~\secref{appendix:training} for additional training details.

To scalably obtain training labels for $\Theta$, we design a distillation label generation module that leverages VFMs to automatically generate training labels. As shown in~\figref{perceptionpipeline}, our label generator only requires sequential SE(3) robot poses and synchronized RGB–point cloud pairs ($o_\text{rgb, 1:t}$, $o_\text{cloud, 1:t}$) to generate the training labels described next.

\begin{figure*}[!t]
\begin{center}
    \centering
	\includegraphics[width=1.0\textwidth]{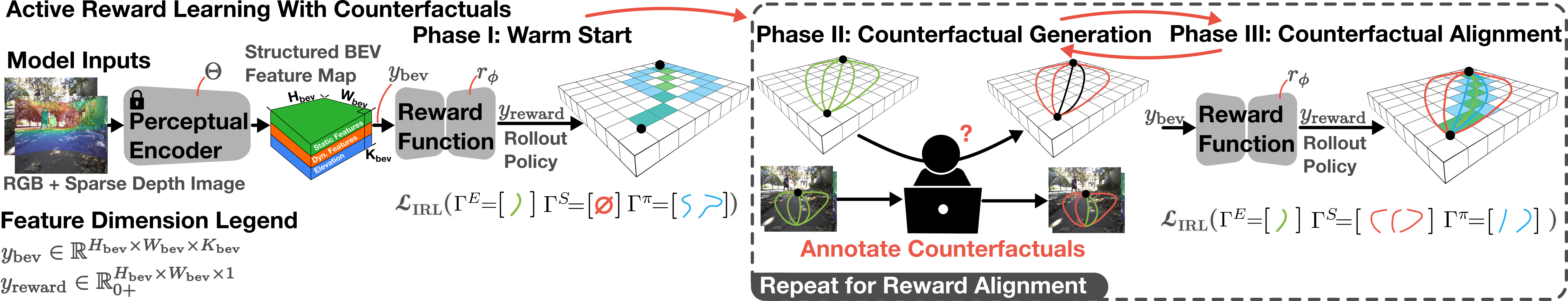}
	\captionof{figure}{Framework for Active Reward Learning with Counterfactuals. Before reward learning, we train and freeze the perceptual encoder $\Theta$ and use the output BEV feature map $y_\text{bev}$ with $r_\phi$ to predict BEV reward maps $y_\text{reward}$. In phase I, we train $r_\phi$ using our counterfactual IRL objective $\mathcal{L}_\text{IRL}$ using only expert demonstrations $\Gamma^E$. In phase II, we plan goal-reaching paths using $y_\text{reward}$ and identify samples that align poorly with human preferences. We generate alternate trajectories for these samples and query the operator to select counterfactual demonstrations $\Gamma^S$ using $o_\text{rgb}$ for context. In phase III, we retrain $r_\phi$ using $\mathcal{L}_\text{IRL}$, this time with $\Gamma^E$ and $\Gamma^S$. We repeat phases II and III to iteratively improve $r_\phi$ until it is aligned with human preferences.}
    \figlabel{rewardframework}
    \vspace{-20pt}
\end{center}%
\end{figure*}

\textbf{i) Label Generation for RGB-D Encoder $f_\text{semantic}$.} We generate training labels for the semantic decoder $f_\text{semantic}$ by passing $o_\text{rgb, t}$ through a frozen Dinov2 encoder and bilinearly interpolating the spatial dimension of our output feature map to match the spatial resolution of $y_\text{semantic}$. We generate depth completion labels $\hat{y}_\text{depth}$ by projecting $o_\text{cloud, t}$ to the image and applying bilateral filtering~\cite{premebida2016high} for edge-aware inpainting, producing a dense depth map that respects object boundaries.

\textbf{ii) Label Generation for BEV Inpainting Backbone $f_\text{bev}$.} To train $f_\text{bev}$, we first generate $\hat{y}_\text{bev, dynamic}$ by prompting SAM2 with bounding boxes of common dynamic classes (e.g., vehicles, pedestrians), backproject the masks to 3D using $o_\text{cloud, t}$, and apply DBSCAN~\cite{ester1996density} clustering at multiple density thresholds. The dynamic BEV map retains clusters with sufficient IoU overlap with the SAM2 instance labels. This procedure mirrors prior works~\cite{ovsep2025better} that leverage SAM2 for learning instance-aware 3D representations. To generate $\hat{y}_\text{bev, static}$, we query SAM2 with a grid of points, filter out overlaps with dynamic masks to isolate static segments (generating dynamic masks as before), and apply a greedy IoU-based merging strategy across frames to maintain instance consistency across frames. The merged static masks are then projected and accumulated in BEV using known robot poses. We refer readers to the Appendix~\secref{appendix:perceptionlabelgeneration} for our algorithmic formulation describing the merging procedure. Lastly, we generate $\hat{y}_\text{bev, elev}$ by accumulating static 3D points across sequential frames using robot poses (generating static masks as before), assign each 3D point to a grid cell, and compute each cell’s minimum elevation by averaging the N lowest 3D points that fall into that cell.
\subsubsection{Training the Reward Function $r_\phi$}
\sseclabel{rewardfunctionlearning}

After freezing $\Theta$, we train $r_\phi$ using our counterfactual-based reward learning objective $\mathcal{L_\text{IRL}}$, where counterfactuals hold all other variables constant with the expert except for the path taken to the end goal. $\mathcal{L_\text{IRL}}$ optimizes the reward function such that the likelihood of counterfactual trajectories is minimized and the likelihood of expert trajectories is maximized, which we observe improves the sample efficiency, expert alignment, and interpretability of the learned rewards. 

To understand $\mathcal{L}_\text{IRL}$, we formally refer to counterfactuals as suboptimal trajectories and define a state-action visitation distribution to be the discounted probability of reaching a state $s$ under a policy $\pi$ and taking action $a$: $\rho^\pi(s,a)=(1-\gamma)\sum \gamma^t p(s_t=s,a_t=a|\pi)$. We define each state $s$ to be an xy location on the local BEV grid and our actions $a$ along the 8 connected grid. Under this formulation, the Counterfactual IRL objective is simple: For each BEV observation $y_{bev}$, given state-actions sampled from the expert's visitation distribution $\rho^E$ and state-actions sampled from the suboptimal visitation distribution $\rho^S$; we obtain a reward function and policy by solving the following optimization problem:
\begin{multline}
    \min_\pi \max_\phi \mathbb{E}_{\rho^E}[r_\phi(s,a,y_{bev})] \\- (\alpha  \mathbb{E}_{\rho^S}[r_\phi(s,a,y_{bev})] + (1-\alpha) \mathbb{E}_{\rho^\pi}[r_\phi(s,a,y_{bev})])
    \eqnlabel{counterfactualirlexpectation}
\end{multline}
where $\rho_\pi$ denotes current policy visitation and $\alpha$ is a tunable hyperparameter that balances the relative importance between suboptimal and expert demonstrations. 

Using the relationship defined in~\eqnref{convex_combination_ranking}, we can rewrite~\eqnref{counterfactualirlexpectation} in terms of the state-action visitation distribution to obtain our Counterfactual IRL training objective:
\begin{equation}
    \mathcal{L_\text{IRL}} = \rho^E(s, a)  - ( \alpha \rho^S(s, a) + (1-\alpha) \mathbb{E}_\pi[\rho^\pi(s, a)] ) r_\phi
\end{equation}
Assuming non-trivial rewards, which can be enforced using gradient regularization techniques~\cite{ma2022smodice}, the above objective learns a reward function such that the difference between the expert's return and agent policy's return is minimized while ensuring suboptimal counterfactuals have a low return. Next, we describe our active learning framework for learning $r_\phi$ from counterfactuals and how we generate these counterfactual annotations. We conclude by deriving the reward learning objective using the Bradley-Terry model of preferences and show connections to Inverse Reinforcement Learning (IRL).

\textbf{Active Reward Learning from Counterfactuals.} Our reward learning framework uses Counterfactual IRL to learn a mapping from BEV features in $y_\text{bev}$ to their scalar utilities. Our framework trains $r_\phi$ in multiple phases, iteratively prompting expert operators for counterfactual annotations in underperforming scenarios until the learned rewards are sufficient for planning trajectories similar to expert demonstrations. \figref{rewardframework} illustrates this framework, which we describe next:

\textit{Phase I: Warmstart} We obtain a base reward function by training $r_\phi$ using only expert demonstrations. This can be done simply by setting $\alpha$ equal to zero in $\mathcal{L}_\text{IRL}$.

\textit{Phase II: Synthetic Counterfactual Generation} We rollout a policy using the learned rewards from phase I and select training samples needing refinement using the Hausdorff distance between the policy rollouts and expert demonstrations. For samples exceeding a distance threshold, we accumulate RGB observations to BEV and generate candidate trajectories sharing start/end points with the expert. A human annotator using our counterfactual annotation tool in~\figref{counterfactualannotationtool} to select trajectories that violate their preferences (e.g., collisions, undesirable terrain) as counterfactuals, which are used in phase III to retrain the reward function using $\mathcal{L}_\text{IRL}$.

\textit{Phase III: Counterfactual Reward Alignment} We retrain $r_\phi$ using $\mathcal{L}_\text{IRL}$ using phase II's counterfactual annotations and the original expert demonstrations.  We set $\alpha$ to be nonzero to balance their relative importance. We repeat phases II and III until the learned policy converges with expert behavior.

\textbf{Counterfactual Generation Details.} Counterfactual IRL relies on being able to sample counterfactual trajectories that visit a diverse set of states between the start and goal. To do this, we propose a simple method of perturbing the expert trajectory to obtain these counterfactuals. Given the expert trajectory $\Gamma^E_t$, we sample a handful of "control" states at regular intervals along $\Gamma^E_t$, excluding the start and goal states. We perturb each control state according to a non-zero mean Gaussian distribution centered at each control state $\mathcal{N}(\mu+s, \sigma)$ before planning a kinematically feasible path that reaches the start, goal, and perturbed control states. Practically, we implement this using Hybrid A*~\cite{dolgov2008practical} - however, any kinematic planner will suffice. We repeat this process twice using positive and negative $\mu$ terms to sample paths ($\sim10$) on both sides of the expert before passing these paths to the operator for counterfactual labeling. For additional implementation details regarding generating alternate trajectories, we refer readers to Appendix~\secref{appendix:counterfactualgeneration}.


\textbf{Counterfactual IRL Derivation $\mathcal{L}_\text{IRL}$.} IRL methods~\cite{ziebart2008maximum,ni2021f} allow for learning reward functions $r_\phi$, parameterized by $\phi$, given expert demonstrations. However, they provide no mechanism to incorporate suboptimal trajectories. Suboptimal trajectories are easy to obtain and enable a data flywheel for navigation; the BEV observations obtained from expert runs can simply be relabeled with suboptimal or unsafe trajectories offline without any more environmental interactions. Motivated by this idea, we derive $\mathcal{L_\text{IRL}}$, a general and principled way to learn from suboptimal and expert trajectories jointly under a single objective. 

Our approach builds on the ranking perspective of imitation learning~\cite{sikchi2022ranking} which uses visitation distributions to denote long-term behavior of an agent. We denote $\rho^\pi(s,a)$, $\rho^E(s,a)$, $\rho^S(s,a)$ to be the agent, expert, and suboptimal state-action visitation distributions respectively. We assume the reward function is conditioned on $y_{bev}$ as before, but drop it from the derivation for conciseness. Under this notation the problem of return maximization becomes finding a visitation induced by a policy $\pi$ that maximizes the expected return given by:
\begin{equation}
    \max_{\pi} J^\pi(r_\phi) = \max_{\pi} \mathbb{E}_{\rho^\pi(s,a)}[r_\phi(s,a)].
\end{equation}
The reward function of the expert should satisfy the ranking $\rho^\pi(s,a) \preceq \rho^E(s,a)$, which implies that the expert's visitation distribution obtains a return that is greater or equal to any other policy's visitation distribution in the environment:
\begin{multline}
     \rho^\pi(s,a) \preceq \rho^E(s,a) \implies \mathbb{E}_{\rho^\pi(s,a)}[r_\phi(s,a)] \\\le \mathbb{E}_{\rho^E(s,a)}[r_\phi(s,a)].
\end{multline}
This property extends to any suboptimal visitations, and as a consequence of linearity of expectations, to any convex combination of the current policy's visitation and any other suboptimal visitation distribution. Mathematically, this is:
\begin{equation}
\label{eq:convex_combination_ranking}
\alpha \rho^\pi(s,a) + (1-\alpha) \rho^S(s,a) \preceq \rho^E(s,a) ~\forall \alpha\in[0,1].
\end{equation}
Thus, given suboptimal visitation distributions, we can create a number of pairwise preferences by choosing a suboptimal visitation and a particular $\alpha$. We turn to the Bradley-Terry model of preferences to satisfy these pairwise preferences which assumes that preferences are noisy-rational and that the probability of a preference can be expressed as:
\begin{equation}
\begin{aligned}
P(& \rho^E(s,a) \succeq \alpha \rho^\pi(s,a) + (1-\alpha) \rho^S(s,a)) = \\
& \frac{e^{J^{E}(r_\phi)}}{e^{J^{E}(r_\phi)} + e^{ \alpha J^{\pi}(r_\phi) + (1 - \alpha)J^{S}(r_\phi)} }\\
&= \frac{1}{1 + e^{\alpha (J^{S}(r_\phi) - J^{E}(r_\phi)) + (1 - \alpha) ( J^{\pi}(r_\phi) - J^{E}(r_\phi)) }}.
\end{aligned}
\eqnlabel{eq9bradley}
\end{equation}

Finding a reward function implies maximizing the likelihood of observed preferences while the policy optimizes the learned reward function. Since, the convex combination holds for all values of $\alpha \in [0,1]$, we consider optimizing against the worst-case to obtain the following two-player counterfactual IRL objective, where the reward player learns to satisfy rankings against the worst-possible $\alpha$: 
\begin{equation}
    \max_\phi \min_\alpha P\left(\rho^E(s,a) \succeq \alpha \rho^\pi(s,a) + (1-\alpha) \rho^S(s,a)\right)
\end{equation}
and the policy player maximizes expected return:
\begin{equation}
    \max_{\pi} J^\pi(r_\phi). 
\end{equation}

In practice, optimizing for worst-case $\alpha$ for each BEV scene $y_{bev}$ can quickly make solving the optimization objective challenging due to the large number of scenes we train the reward function on. We make two mild approximations that we observed to make learning more efficient and tractable: First, we replace the worst-case $\alpha$ with a fixed $\alpha$, and second, we consider maximizing a pointwise monotonic transformation to the Bradley Terry loss function that directly maximizes $\alpha (J^{S}(r_\phi) - J^{E}(r_\phi)) + (1 - \alpha) ( J^{\pi}(r_\phi) - J^{E}(r_\phi)) $ instead of its sigmoid transformation. With these changes, we can rewrite our practical counterfactual IRL objective as:

\begin{equation}
    \min_\pi \max_\phi  \left( J^{E}(r_\phi) - (\alpha J^{S}(r_\phi) + (1 - \alpha) J^{\pi} (r_\phi)) \right).
\end{equation}

This objective reveals a deeper connection between apprenticeship learning (Eq 6~\cite{abbeel2004apprenticeship}) obtained by setting $\alpha$ to 0 and learning from preferences~\cite{brown2019extrapolating} obtained by setting $\alpha$ to 1. The loss function goes beyond the apprenticeship learning objective that only learns from expert by incorporating suboptimal demonstrations. Second, it goes beyond the offline nature of prior algorithms that learn from preferences alone by instead learning a policy that attempts to match expert visitation making use of the suboptimal demonstrations.

\section{Implementation Details}
\seclabel{implementation}

In this section, we cover implementation details for our local planning and control modules depicted in~\figref{architectureoverview}. For additional information regarding model training hyperparameters and generating counterfactual annotations, please refer to Appendix~\secref{appendix}.

\subsubsection{Global and Local Planning and Controls}
\sseclabel{implementation:planning}
Our global planning module plans a global path and identifies the next local subgoal for our local planner to plan local paths. Given a user-specified GPS end goal $G_N$, we use OpenStreetMap~\cite{OpenStreetMap} to obtain a semi-dense sequence of coarse GPS goals $G = [G_1, ..., G_N]$ spaced 10 meters apart. To select the next GPS subgoal, we compute the set of distances $D = \{ ||g - G_N|| \text{ } |\text{ } g \in G \cup \{G_t\} \}$, where $D$ contains the distance of the robot $G_t$ from $G_N$ and the distance of each GPS subgoal in $G$ from $G_N$. From the set of subgoals with a smaller distance to $G_N$ than $G_t$, we select the farthest subgoal to use as the next subgoal. We project this subgoal on the edge of the local planning horizon, a 6 meter circle around the robot, giving us a carrot for local path planning.

We adopt a DWA~\cite{fox1997dynamic} style approach to local path planning, where we enumerate a set of constant curvature arcs (31 in our case) from the egocentric robot frame. We compute a scalar cost for each trajectory by sampling points along each each arc and computing the discounted cost at each point using our predicted reward map. We simply invert and normalize our reward map to the range $[0, 1]$ to convert it to a costmap. Finally, we compute the distance between the last point on the arc with the local carrot to obtain a goal-reaching cost. We multiply the learned and goal-reaching costs by tunable weights before selecting the trajectory with the lowest cost. Finally, we perform 1D time optimal control~\cite{pontryagin2018mathematical} to generate low-level actions to follow this trajectory.  Empirically, we find that sampling 30 points and using a discount factor of 0.95 is sufficiently dense. We find that tuning the goal-reaching cost to be 1/10th the importance of the learned cost achieves good balance between goal-reaching and adhering to operator preferences.

\section{experiments}
\seclabel{experiments}

\begin{figure*}[!htp]
\begin{center}
    \centering
	\includegraphics[width=1.0\linewidth]{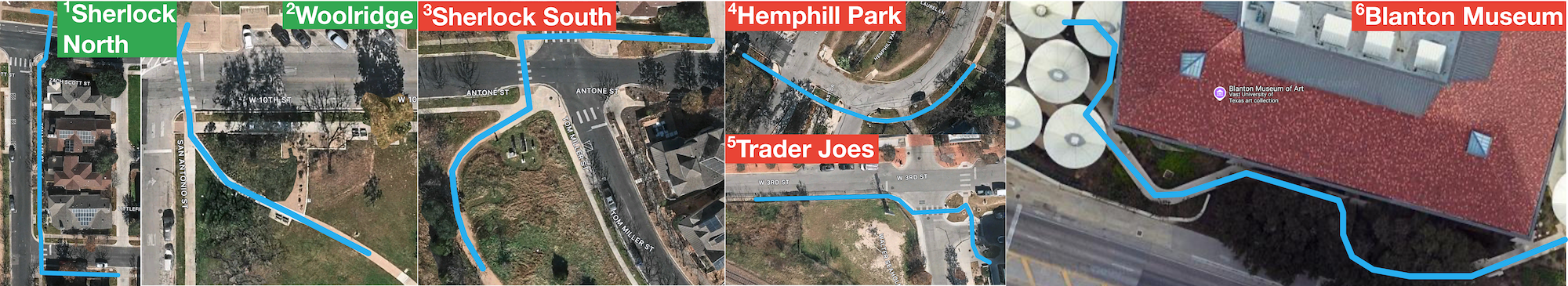}
	\captionof{figure}{Satellite image of testing locations for short horizon mapless navigation experiments. We evaluate baselines across 2 seen (green) and 4 unseen (red) urban locations. Our testing locations consist of residential neighborhoods, urban shopping centers, urban parks, and offroad trails. In each location, each baseline must start from an endpoint on the annotated blue trajectory and navigate to the opposite end of the trajectory. We denote each location's ID with a numerical superscript.}
    \figlabel{shorthorizonexperiments}
    \vspace{-10pt}
\end{center}
\end{figure*}
\begin{figure*}[!t]
\begin{center}
    \centering
	\includegraphics[width=1.0\linewidth]{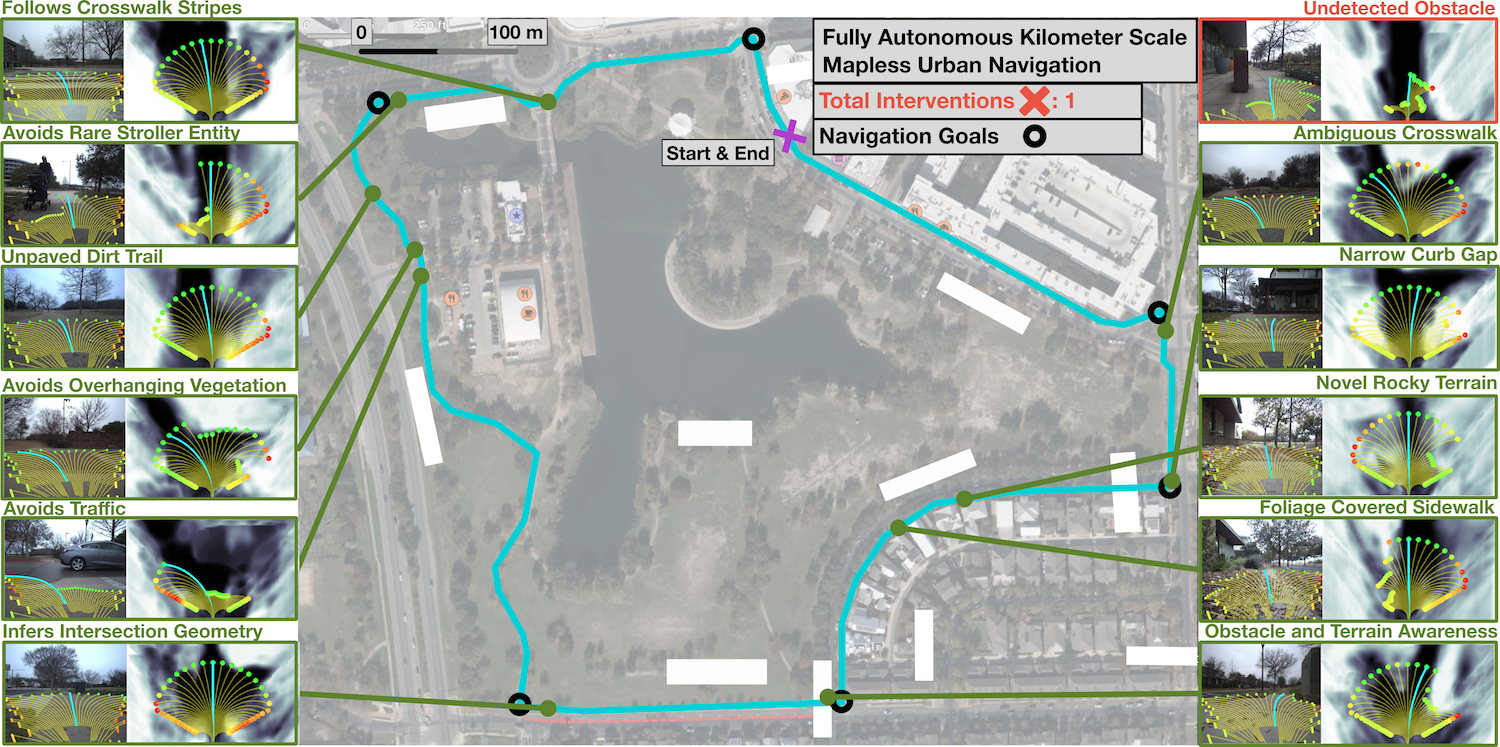}
	\captionof{figure}{Satellite image of our 2 kilometer long-horizon testing area, with examples with front view RGB image observations and \ourmodel{}'s predicted BEV costmap (converted from the predicted BEV reward map). We annotate successful examples in green with a brief description of the situation. We annotate unsuccessful examples in red, present the observation right before the intervention, and provide a brief description of the cause of failure. }
    \figlabel{longhorizonexperiments}
    \vspace{-20pt}
\end{center}%
\end{figure*}

In this section, we describe our evaluation methodology for \ourmodel{} and answer the following questions to understand the importance of our contributions and overall performance on the task of mapless urban navigation.

\begin{itemize}[]
    \item (\textit{$Q_1$}) How well does \ourmodel{} generalize to unseen urban environments for mapless urban navigation?
    \item (\textit{$Q_2$}) How important are structured BEV perceptual representations for downstream policy learning?
    \item (\textit{$Q_3$}) How much do counterfactual demonstrations improve urban navigation performance?
    \item (\textit{$Q_4$}) How well does \ourmodel{} perform long horizon mapless urban navigation compared to other top state-of-the-art approaches?
\end{itemize}

\begin{wrapfigure}{l}{0.2\textwidth}
    \centering
    \vspace{-10pt}
    \includegraphics[width=\linewidth]{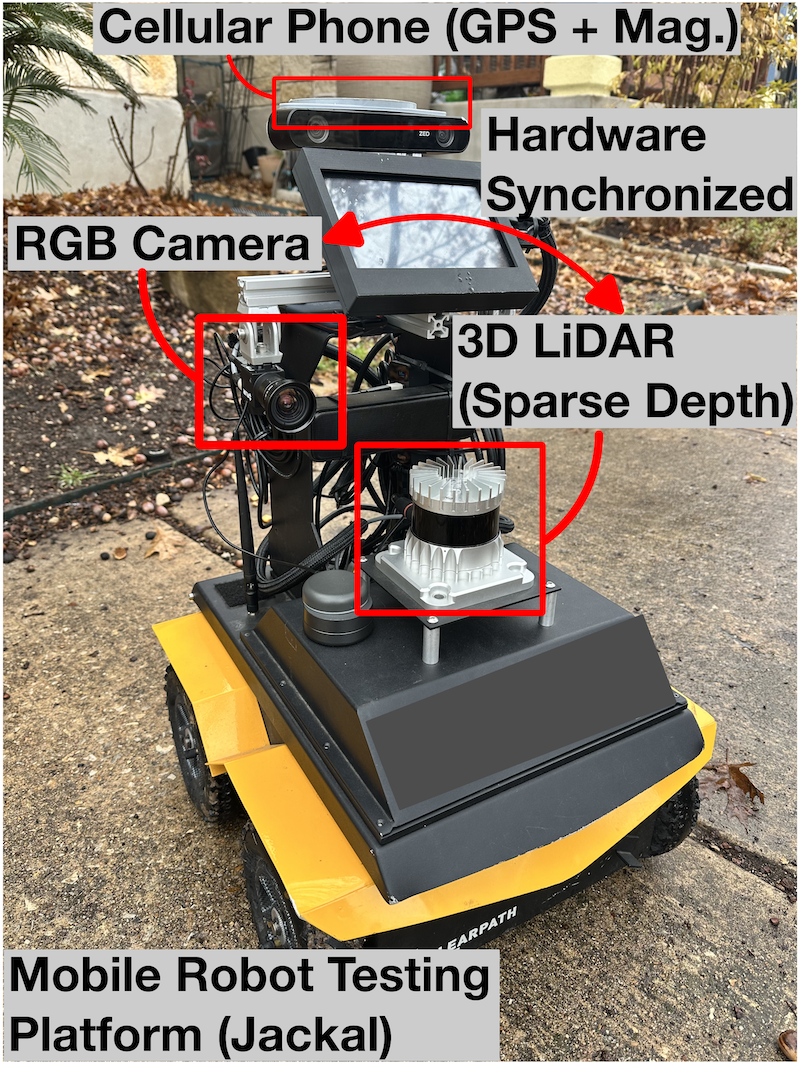}
    \caption{Mobile robot testing platform for real-world experiments. We annotate the locations of the monocular camera, 3D LiDAR, and cellular phone (not visible) on our testing platform, the Clearpath Jackal.}
    \label{robothardware}
    \vspace{-20pt}
\end{wrapfigure}
We investigate $\mathcal{Q}_1$ by comparing \ourmodel{}'s performance against other methods in unseen environments. Additionally, we evaluate the relative performance between different input modalities and observation encoders for \ourmodel{}. To answer $\mathcal{Q}_2$ and $\mathcal{Q}_3$, we conduct ablation studies that isolate the methodological contribution in question. Finally, we evaluate $\mathcal{Q}_4$ by conducting a kilometer-scale experiment comparing \ourmodel{} against the top-performing baseline.

\begin{table*}[]
\begin{tabular}{|l|llllll|lllllllll|}
\hline
& \multicolumn{6}{l|}{\textbf{In Distribution}} & \multicolumn{9}{l|}{\textbf{Out of Distribution}}\\ \hline
\textbf{Location} & \multicolumn{3}{l|}{Sherlock North} & \multicolumn{3}{l|}{Woolridge Square} & \multicolumn{3}{l|}{Sherlock South} & \multicolumn{3}{l|}{Trader Joes} & \multicolumn{3}{l|}{Hemphill Park}\\ \hline
\textbf{Method} & \multicolumn{1}{|l|}{\makecell{AST \(\downarrow\)}} & \multicolumn{1}{|l|}{\makecell{\%S\(\uparrow\)}} & \multicolumn{1}{|l|}{\makecell{NIR \(\downarrow\)}} & \multicolumn{1}{|l|}{\makecell{AST \(\downarrow\)}} & \multicolumn{1}{|l|}{\makecell{\%S\(\uparrow\)}} & \multicolumn{1}{|l|}{\makecell{NIR \(\downarrow\)}} & \multicolumn{1}{|l|}{\makecell{AST \(\downarrow\)}} & \multicolumn{1}{|l|}{\makecell{\%S\(\uparrow\)}} & \multicolumn{1}{|l|}{\makecell{NIR \(\downarrow\)}} & \multicolumn{1}{|l|}{\makecell{AST \(\downarrow\)}} & \multicolumn{1}{|l|}{\makecell{\%S\(\uparrow\)}} & \multicolumn{1}{|l|}{\makecell{NIR \(\downarrow\)}} & \multicolumn{1}{|l|}{\makecell{AST \(\downarrow\)}} & \multicolumn{1}{|l|}{\makecell{\%S\(\uparrow\)}} & \multicolumn{1}{|l|}{\makecell{NIR \(\downarrow\)}} \\ \hline
Geometric Only & \mc{17.64} & \mc{61.60} & \mc{11.21} & \mc{18.56} & \mc{86.36} & \mc{9.50} & \mc{15.56} & \mc{92.50} & \mc{7.80} & \mc{15.22} & \mc{94.74} & \mc{12.05} & \mc{14.21} & \mc{95.00} & \mc{13.21}\\ \hline
PACER+G~\cite{mao2024pacer} & \mc{25.79} & \mc{96.15} & \mc{9.42} & \mc{26.51} & \mc{90.90} & \mc{8.83} & \mc{22.11} & \mc{95.00} & \mc{7.67} & \mc{24.38} & \mc{87.14} & \mc{17.93} & \mc{23.94} & \mc{92.85} & \mc{9.47} \\ \hline
ViNT~\cite{shah2023vint} & \mc{20.38} & \mc{65.51} & \mc{10.36} & \mc{17.85} & \mc{90.36} & \mc{7.94} & \mc{17.06} & \mc{92.30} & \mc{7.19} & \mc{22.10} & \mc{84.21} & \mc{18.36} & \mc{23.92} & \mc{95.00} & \mc{10.96} \\ \hline
PIVOT~\cite{nasiriany2024pivot} & \mc{46.56} & \mc{83.33} & \mc{26.03} & \mc{57.36} & \mc{84.11} & \mc{20.22} & \mc{45.41} & \mc{75.00} & \mc{21.91} & \mc{41.77} & \mc{74.44} & \mc{40.67} & \mc{33.85} & \mc{85.00} & \mc{28.36} \\ \hline
\ourmodel{} - cfs - st & \mc{21.26} & \mc{96.15} & \mc{12.99} & \mc{20.53} & \mc{91.66} & \mc{9.79} & \mc{21.13} & \mc{92.80} & \mc{5.65} & \mc{23.17} & \mc{93.75} & \mc{14.87} & \mc{26.85} & \mc{94.11} & \mc{11.11}\\ \hline
\ourmodel{} - cfs  & \mc{25.86} & \mc{96.29} & \mc{9.20} & \mc{19.84} & \mc{90.90} & \mc{6.12} & \mc{13.27} & \mc{92.30} & \mc{2.41} & \mc{25.39} & \mc{91.66} & \mc{10.49}  & \mc{28.43} & \mc{83.33} & \mc{8.57} \\ \hline
\ourmodel{} - st  & \mc{19.09} & \mc{96.29} & \mc{6.10} & \mc{17.80} & \mc{92.90} & \mc{3.03} & \mc{12.88} & \mc{94.65} & \mc{2.04} & \mc{22.01} & \mc{92.85} & \mc{7.83}  & \mc{16.32} & \mc{93.42} & \mc{2.16} \\ \hline
\ourmodel{} (ours) & \mc{\textbf{14.01}} & \mc{\textbf{96.60}} & \mc{\textbf{4.60}} & \mc{\textbf{14.70}} & \mc{\textbf{100.0}} & \mc{\textbf{0.00}} & \mc{\textbf{12.60}} & \mc{\textbf{95.23}} & \mc{\textbf{0.86}} & \mc{\textbf{15.28}} & \mc{\textbf{95.65}} & \mc{\textbf{3.73}} & \mc{\textbf{15.42}} & \mc{\textbf{100.0}} & \mc{\textbf{0.00}} \\ \hline
\end{tabular}
\caption{Quantitative evaluation comparing \ourmodel{} against existing navigation baselines for short horizon mapless navigation experiments. In distribution locations are present in the training dataset while out-of-distribution locations are absent from the training data. We bold the best performing method for each metric in each location, and annotate each metric with an up or down arrow to indicate if higher or lower numbers are better. We define the following evaluation metrics in the evaluation metric section of this work and denote their abbreviations as: AST - average subgoal completion time (s), \%S - percentage of subgoals reached in mission, NIR - Number of interventions required per 100 meters driven.}
\tablabel{shorthorizon}
\vspace{-20pt}
\end{table*}

\begin{table}[]
\begin{tabular}{|l|lll|}
\hline
& \multicolumn{3}{l|}{\textbf{Out of Distribution}} \\ \hline
\textbf{Location} & \multicolumn{3}{l|}{ Blanton Museum}\\ \hline
\textbf{Method} & \multicolumn{1}{|l|}{\makecell{AST \(\downarrow\)}} & \multicolumn{1}{|l|}{\makecell{\%S\(\uparrow\)}} & \multicolumn{1}{|l|}{\makecell{NIR \(\downarrow\)}} \\ \hline
\ourmodel{}-RGB-FROZEN & \mc{17.14$\pm$2.51} & \mc{97.91$\pm$2.59} & \mc{1.69$\pm$0.96} \\ \hline
\ourmodel{}-RGB & \mc{17.25$\pm$3.83} & \mc{97.38$\pm$2.43} & \mc{1.38$\pm$0.90} \\ \hline
\ourmodel{}-STEREO& \mc{17.72$\pm$1.32} & \mc{97.43$\pm$1.88} & \mc{1.07$\pm$0.49} \\ \hline
\ourmodel{} (ours) & \mc{\textbf{13.81$\pm$2.38}} & \mc{\textbf{98.36$\pm$1.55}} & \mc{\textbf{0.76$\pm$0.48}}\\ \hline
\end{tabular}
\caption{Quantitative evaluation comparing the impact of different input modalities and observation encoders on \ourmodel{}. All baselines are evaluated on the same experiment area. We compute the mean and standard deviation over 5 trials and bold the best performing method for each metric in each location. We annotate each metric with an up or down arrow to indicate if higher or lower numbers are better. 
}
\tablabel{shorthorizonmodality}
\vspace{-15pt}
\end{table}

\subsection{Robot Testing Platform}

We conduct all experiments using a Clearpath Jackal mobile robot. We observe $512 \times 612$ RGB images from a $110^{\circ}$ field-of-view camera and point cloud observations from a 128-channel Ouster LiDAR. We obtain coarse GPS measurements and magnetometer readings from a cellular smartphone. We use the open source OpenStreetMap~\cite{OpenStreetMap} routing service to obtain coarse navigation waypoints. Our onboard compute platform has an Intel i7-9700TE 1.80 GHz CPU and Nvidia RTX A2000 GPU. We run \ourmodel{} at 20 Hz alongside our mapless navigation system, which operates at 10Hz.

\subsection{Training Dataset}

We collect a robot dataset with 3 hours of expert navigation demonstrations spread across urban parks, downtown centers, residential neighborhoods, and college campuses. Additionally, we annotate 3\% of our training dataset with negative counterfactual annotations to train \ourmodel{}. Our dataset consists of synchronized image LiDAR observation pairs and ground truth robot poses computed using LeGO-LOAM~\cite{legoloam2018}, a LiDAR-based SLAM algorithm. We train all methods on the same dataset for 150 epochs or until convergence.

\subsection{Testing Methodology}
We evaluate all questions through physical robot experiments performing the task of mapless urban navigation in urban environments. We present aerial images in~\figref{shorthorizonexperiments} depicting the urban environments where we perform short horizon ($\sim$100m) quantitative experiments. These consist of six challenging seen and unseen environments with trails, road crossings, narrow pathways, different terrains, and obstacle hazards. For locations 1-5, we repeat the same experiment twice for each approach. For location 6, we repeat the same experiment five times for each approach. We evaluate the highest performing methods on a long-horizon quantitative experiment, which we conduct at a 2-kilometer urban trail shown in~\figref{longhorizonexperiments}. We pre-emptively terminate the long-horizon experiment if the baseline is unable to complete the mission within a predefined time. Next, we explain evaluation metrics 1 to 3, which we use for the short-horizon experiments, and evaluation metrics 4 and 5, which we use exclusively for the long-horizon experiments.

Our evaluation metrics are defined as follows: 1) \textit{Average Subgoal Completion Time (AST)} - the average time to complete each subgoal where all subgoals are evenly spaced 10 meters apart 2) \textit{Percentage of Subgoals Reached (\%S)} - the percentage of subgoals reached by the end of the mission 3) \textit{Normalized Intervention Rate (NIR)} - the number of operator interventions required for every 100 meters driven 4) \textit{Total Distance Driven (Dist. (m))} - the total distance driven before mission failure or completion 5) \textit{Total Interventions (Total Int.)} - the total number of interventions required per mission. During evaluation, we only consider interventions that were incurred by the method (e.g. overrides performed to give right of way to vehicular traffic are not considered interventions).

\subsection{Baselines}
\seclabel{experiments:baselines}
 We supplement all baselines with the same analytical geometric avoidance module to prevent catastrophic collisions. Even with this module, operators preemptively intervene when the baseline deviates from general navigation preferences (e.g. stay on sidewalks, follow crosswalk markings, etc). All baselines, with the exception of PIVOT~\cite{nasiriany2024pivot}, perform all computations using onboard compute to fairly evaluate real-time performance. Next, we describe our evaluation baselines.

For questions 1 and 3, we compare \ourmodel{} against four existing navigation baselines and six variations of the \ourmodel{} architecture. We first describe the existing baselines: 1) \textit{ViNT}~\cite{shah2023vint} - a foundation navigation model trained on goal-guided navigation 2) \textit{PACER+Geometric}~\cite{mao2024pacer} (PACER+G) - a multi-factor perception baseline that considers learned terrain and geometric costs in a dynamic window~\cite{fox1997dynamic} (DWA) style approach 3) \textit{Geometric-Only} - a DWA style approach that only considers geometric costs 4) \textit{PIVOT}~\cite{nasiriany2024pivot} - a VLM-based navigation method that uses the latest version of GPT4o-mini~\cite{achiam2023gpt} to select local goals from image observations. 

We test six variations of \ourmodel{} to evaluate the effect of different observation encoders and sensor modalities. These consist of 1) \ourmodel{}-RGB, the same RGB encoder but using only monocular RGB images, 2) \ourmodel{}-STEREO, the same RGB encoder with a stereo processing backbone that fuses features from stereo RGB images, 3) \ourmodel{}-RGB-Frozen, the same depth prediction head but with a frozen Dinov2~\cite{oquab2023dinov2} encoder that only uses monocular RGB images. The remaining variations ablate our methodological contributions:  4) \ourmodel{}-cfs, our model trained without counterfactual demonstrations, but otherwise identical to the original, 5) \ourmodel{}-st, our model trained without the BEV inpainting backbone, but otherwise identical to the original, 6) \ourmodel-cfs-st, our model trained without counterfactual demonstrations or the BEV inpainting backbone. For architectures 5 and 6, we directly pass the unstructured BEV feature map $z_\text{bev, splat}$ to the reward function $r_\phi$ and train the splat module $f_\text{splat}$ using expert demonstrations. We still freeze the RGB-D backbone $f_\text{rgbd}$ when training $r_\phi$. For additional details regarding the \ourmodel{} architecture variations, we refer readers to Appendix~\ssecref{appendix:failureanalysismodality}. Next, we will describe high-level implementation details for each non-\ourmodel{} baseline.

Since ViNT is pre-trained for image goal navigation, we are unable to deploy this model in unseen environments where image goals are not known apriori. Thus, we finetune ViNT by freezing the pre-trained ViNT backbone and changing the goal modality encoder to accept 2D xy coordinate goals rather than images. We follow the same model architecture and match the training procedure from the original paper for reaching GPS goals. We reproduce the PACER and PIVOT models faithfully to the best of our abilities as no open-source implementation is available. 

\subsection{Quantitative Results and Analysis}
\seclabel{quantitativeanalysis}

\begin{table}[]
\begin{tabular}{|l|lllll|}
\hline
\textbf{Location} & \multicolumn{5}{l|}{Mueller Loop}\\ \hline
\textbf{Method} & \multicolumn{1}{|l|}{\makecell{AST \(\downarrow\)}} & \multicolumn{1}{|l|}{\makecell{\%S\(\uparrow\)}} & \multicolumn{1}{|l|}{\makecell{NIR \(\downarrow\)}} & \multicolumn{1}{|l|}{\makecell{Dist. (m) \(\uparrow\)}} & \multicolumn{1}{|l|}{\makecell{Total Int. \(\downarrow\)}}\\ \hline
PACER+G~\cite{mao2024pacer} & \mc{15.8} & \mc{61.53} & \mc{3.56} & \mc{1345.07} & \mc{48}\\ \hline
\ourmodel{} (ours) & \mc{\textbf{12.94}} & \mc{\textbf{99.45}} & \mc{\textbf{0.052
}} & \mc{\textbf{1919.44}} & \mc{\textbf{1}} \\ \hline
\end{tabular}
\caption{Quantitative evaluation for long horizon mapless navigation experiments. All baselines are evaluated on the same 1.9 kilometer urban area. We bold the best performing method for each metric in each location, and annotate each metric with an up or down arrow to indicate if higher or lower numbers are better. We define the following evaluation metrics in the evaluation metric section of this work and denote their abbreviations as: AST - average subgoal completion time (s), \%S - percentage of subgoals reached in mission, NIR - Number of interventions required per 100 meters driven, Dist. (m) - total distance driven in meters, Total Int. - total number of interventions required for the entire mission.}
\tablabel{longhorizon}
\vspace{-20pt}
\end{table}

We present the quantitative results of the short horizon and long-horizon experiments in \tabref{shorthorizon}, \tabref{shorthorizonmodality}, and \tabref{longhorizon} respectively, and analyze these results to answer $\mathcal{Q}_1$ - $\mathcal{Q}_4$ in the following section.

\subsubsection{Evaluating Generalizability to Unseen Urban Environments.} Comparing \ourmodel{} against all other baselines in \tabref{shorthorizon}, we find that our approach achieves superior performance in all metrics for both seen and unseen environments. We present qualitative analysis for each learned baseline in Appendix~\secref{appendix:shorthorizonanalysis}. Quantitatively, PACER+G, the next best approach, requires two times more interventions than \ourmodel{} in seen environments and five times more interventions in unseen environments. Furthermore, we find that PIVOT, our VLM-based navigation baseline, performs poorly relative to other methods, corroborating our claim that VLMs and LLMs are not well attuned for urban navigation despite containing internet-scale priors. We hypothesize this is because navigation requires identifying which priors are most important for navigation, a task seen rarely by VLMs and LLMs during pre-training. Notably, we achieve an intervention-free traversal in one unseen environment (Hemphill Park), a residential park with diverse terrains, narrow curb gaps, and crosswalk markings. From these findings, we conclude that \ourmodel{} is remarkably more generalizable for mapless urban navigation than existing approaches.

In \tabref{shorthorizonmodality}, our approach performs most favorably using RGB and LiDAR inputs, followed by stereo RGB and monocular RGB respectively. This occurs because LiDAR offers robustness to severe lighting artifacts like lens flares that affect costmap prediction quality. We present qualitative examples of this in~\figref{failureanalysismodality}. Interestingly, we find that distilled Dinov2 features perform favorably compared to frozen Dinov2 features. Prior works corroborate this observation~\cite{yang2023emernerf, yang2024denoising} and verify that Dinov2 features contain positional embedding artifacts that disrupt multi-view consistency. We postulate that our distillation objective enables our RGB-D encoder $f_\text{rgbd}$ to focus on learning artifact-free features while the semantic decoder $f_\text{semantic}$ learns how to reconstruct these artifacts. For additional analysis comparing these architectural variations, we refer readers to Appendix~\ssecref{appendix:failureanalysisbaselines}.

\subsubsection{Evaluating the Importance of Structured BEV Perceptual Representations.} We assess the impact of structured perceptual representations by evaluating \ourmodel{}-cfs-st and \ourmodel{}-st against \ourmodel{} in~\tabref{shorthorizon}, where the only difference is whether structured representations and counterfactuals are used. Without structured representations, \ourmodel{}-st incurs 28\% more interventions on average across all environments. Performance further deteriorates without structured representations or counterfactuals, incurring 41\% more interventions on average across all environments. We hypothesize that this occurs because structured representations encode higher-level features that are more generalizable and easier for policies to reason about compared to lower-level features that capture less generalizable high-frequency information.

\subsubsection{Evaluating the Importance of Counterfactual Demonstrations.} We compare \ourmodel{} against \ourmodel{}-cfs, the exact same approach but without using counterfactuals to train the reward function. On average, we find in~\tabref{shorthorizon} that using counterfactuals reduces the number of interventions by 70\% in seen environments and 69\% in unseen environments. Both of these baselines distill the same features for navigation from VFMs, 

\subsubsection{Evaluating Performance on Kilometer-Scale Mapless Navigation.} \tabref{longhorizon} compares \ourmodel{} against the top-performing baseline from \tabref{shorthorizon}, PACER+G, on long horizon mapless navigation.~\figref{longhorizonexperiments} show qualitative examples of \ourmodel{}'s predicted BEV costmaps along the route. While PACER+G still drives over a kilometer before timing out, it requires significantly more interventions to do so. We observe that the majority of PACER+G failures that occur result from either not adequately considering geometric factors like curb gaps or predicting poor costmaps due to differences in lighting and weather compared to the conditions from which the training dataset was collected. Contrastingly, \ourmodel{} is robust to perceptual aliasing from lighting and weather variations and predicts accurate costmaps even during failures. From this, we conclude that \ourmodel{} achieves superior performance in long-horizon mapless urban navigation as well. 

\section{Limitations and Future Work}
\seclabel{Limitations}

While \ourmodel{} inherits some viewpoint-invariance from VFMs, it does not allow reward maps to be conditioned on embodiment-specific constraints, such as the traversability differences between quadrupeds and wheeled robots. An interesting future direction is applying existing research on cross-embodiment conditioning~\cite{shah2023vint} to our architecture. Furthermore, \ourmodel{} infers reward maps using observations from a single timestep, limiting multi-step horizon reasoning tasks, such as recovering from dead ends or negotiating paths in constricted environments. Extending \ourmodel{} with memory and counterfactual IRL for reasoning about object dynamics are promising directions for addressing these issues. In addition, reward learning with counterfactuals is a potential bottleneck, as it requires humans for counterfactual labeling. Automated counterfactual generation with VLMs/LLMs is another promising direction to further improve scalability. Lastly, our counterfactual IRL objective can further be extended to incorporate preferences following the same derivation from the ranking perspective. This would not only allow the reward to reason about what paths are best but allow is to learn how to discriminate between two suboptimal paths should the need arise.


\section{Conclusion}
\seclabel{conclusion}

In this paper, we introduce Counterfactuals for Reward Enhancement with Structured Embeddings (\ourmodel{}), a scalable framework for learning representations and policies that address the open-set generalization and robustness challenges of open-world mapless navigation. \ourmodel{} learns robust, generalizable perceptual representations with internet-scale semantic, geometric, and entity priors by distilling features from multiple visual foundation models. We demonstrate that our perceptual representation encodes a sufficient set of factors for urban navigation, and introduce a novel counterfactual-based loss and active learning framework that teaches policies to hone in on the most important factors and infer how they influence fine-grained navigation behavior. These contributions culminate to form \ourmodel{}, a local path planning module that significantly outperforms state-of-the-art alternatives on the task of long-horizon mapless urban navigation. Through kilometer-scale real-world robot experiments, we demonstrate our approach's effectiveness, gracefully navigating an unseen 2 kilometer urban environment with only a handful of interventions. 

\section{Acknowledgements}
\seclabel{acknowledgements}
This work has taken place in the Autonomous Mobile Robotics Laboratory (AMRL) and Machine Decision-making through Interaction Laboratory (MIDI) at UT Austin. AMRL research is supported in part by NSF (CAREER-2046955, PARTNER-2402650) and ARO (W911NF-24-2-0025). MIDI research is supported in part by NSF (CAREER-2340651, PARTNER-2402650), DARPA (HR00112490431), and ARO (W911NF-24-1-0193). Any opinions, findings, and conclusions expressed in this material are those of the authors and do not necessarily reflect the views of the sponsors. 


\bibliographystyle{plainnat}
\bibliography{references}

\clearpage
\section{Appendix}
\seclabel{appendix}

We organize the appendix into the following sections: 
1) Details for training \ourmodel{} and other baselines in~\secref{appendix:training}, 2) Details regarding generating counterfactual annotations in~\secref{appendix:counterfactualgeneration}, and 3) Qualitative analysis for the short horizon experiments in~\secref{appendix:shorthorizonanalysis}.

\subsection{Perceptual Encoder Label Generation Details}
\seclabel{appendix:perceptionlabelgeneration}

In this section, we provide additional details regarding generating BEV instance maps for training the perceptual encoder $\Theta$. We observe that the default SAM2 model settings perform well for generating instance labels without significant hyperparameter tuning. When merging sequential SAM2 instance labels to generate static BEV instance maps, it is important to ensure that instance IDs remain temporally consistent when merging cross-view instances. We ensure this by greedily merging BEV instances between frames with the highest intersection over union (IoU). Instances that do not exceed an IoU of 0.2 are treated as unique instances and are merged into the aggregated BEV instance map as such. We repeat this procedure from the first to the last observation in the horizon.

When generating dynamic BEV instance maps, we backproject the image instance labels to 3D and cluster the point cloud using DBSCAN with three density thresholds (0.1, 0.2, 0.3) and require a minimum of (5, 3 5) points in each cluster, respectively. We observe that these thresholds work well across environments for identifying clusters within a 12.8m sensing range. To match SAM2 instance labels to cluster IDs, we use an IoU threshold of 0.2 to determine if the label and cluster IDs should be matched. All unmatched clusters are discarded and the matched clusters are used to construct the dynamic BEV instance map.

\subsection{Model Training Details}
\seclabel{appendix:training}

\begin{figure}[!t]
    \centering
    \vspace{-10pt}
    \includegraphics[width=0.8\linewidth]{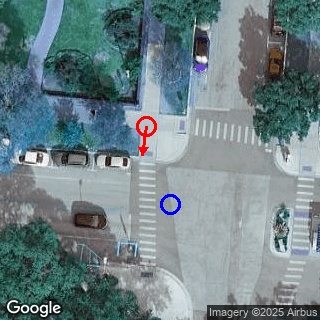}
    \includegraphics[width=\linewidth]{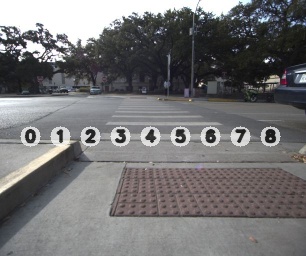}
    \caption{Qualitative Example of context images provided to PIVOT~\cite{nasiriany2024pivot}, our VLM-based navigation baseline. We annotate satellite images with the robot GPS and heading (red) and current GPS goal (blue). Additionally, we annotate the front view image with potential subgoals from which we prompt the model to choose from. }
    \figlabel{appendix:pivotcontext}
    \vspace{-10pt}
\end{figure}

In this section, we supplement the implementation details for \ourmodel{} and each baseline described in~\secref{experiments:baselines}.

\subsubsection{\ourmodel{}} We provide specific architecture and training hyperparameters settings in~\tabref{modelhyperparameters} and \tabref{traininghyperparameters}, and supplement the \ourmodel{} training procedure in~\ssecref{perceptiontrainingprocedure} with additional details for training our observation encoder $\Theta$. We warm-start the RGB-D backbone for 50 epochs or until convergence. We set $\alpha_1$ and $\alpha_2$, the loss weights for semantic feature regression and completion, to 1 and 0.5 respectively and keep this fixed for the rest of training. After this, we freeze the RGB-D backbone and train the $f_\text{bev}$ using the BEV inpainting backbone loss $\mathcal{L}_\text{bev}$ for five epochs before unfreezing the RGB-D backbone and training end-to-end for another 45 epochs or until convergence. This enables faster convergence by stabilizing $f_\text{bev}$ before joint training. We set $\beta_1$, $\beta_2$, and $\beta_3$ to 1, 2, and 3 for $\mathcal{L}_\text{bev}$, but find that training remains stable for any reasonable combination of values. Finally, we empirically find that replacing the Supervised Contrastive loss~\cite{khosla2020supervised} with Cross Entropy loss while training the dynamic panoptic head $f_\text{bev, dynamic}$ empirically improves overall performance. Thus, we use this model for our long-horizon navigation experiments presented in~\tabref{longhorizon}.

As mentioned in~\ssecref{rewardfunctionlearning}, we train our reward function $r_\phi$ in two phases. During both phases, we freeze $\Theta$, thus only training $r_\phi$. During initialization (Phase I), we train for 25 epochs, setting $\alpha=0$ only to use expert demonstrations. During finetuning (Phase III), we initialize $r_\phi$ from scratch, train for 50 epochs, setting $\alpha=0.5$ and a reward regularization term~\cite{ma2022smodice} $\alpha_{reg} = 1.0$. Empirically, we find that our objective is stable for a range of $\alpha$ values and benefits from a larger reward regularization penalty to encourage more discriminative rewards. In total, we train the perceptual encoder $\Theta$ and reward function $r_\phi$ for a combined 150 epochs.

\subsubsection{ViNT~\cite{shah2023vint}} To modify ViNT to follow XY goal guidance instead of image goal guidance, we follow the architecture design details in the ViNT paper for adapting the backbone to GPS goal guidance. We only train the XY goal encoder and freeze the remaining model parameters. We sample XY goals from future odometry between 8 to 10 seconds to generate training data.

\subsubsection{PIVOT~\cite{nasiriany2024pivot}} We condition PIVOT using an annotated satellite view image, annotated front view image, and text instructions. As shown in~\figref{appendix:pivotcontext}, we annotate a satellite image containing the robot's current position, heading, and next goal GPS coordinates.~\figref{appendix:pivotcontext} also presents the annotated front view RGB image with each numbered circle indicating a potential goal that our VLM must choose from. Finally, we provide the text prompt below:
\begin{quote}
I am a wheeled robot that cannot go over objects. This is the image I'm seeing right now. I have annotated it with numbered circles. Each number represents a general direction I can follow. I have annotated a satellite image with my current location and direction as a red circle and arrow and the goal location as a blue circle. 
Now you are a five-time world champion navigation agent and your task is to tell me which circle I should pick for the task of: going forward \textbf{X} degrees to the \textbf{Y}?
Choose the best candidate number. Do NOT choose routes that go through objects AND STAY AS FAR AS POSSIBLE AWAY from untraversable terrains and regions. Skip analysis and provide your answer at the end in a json file of this form:
{{"points": [] }}
\end{quote}
where X is replaced by the degrees from the goal computed using the magnetometer heading and goal GPS location, and Y is either left or right depending on the direction of the goal. During testing, we use our geometric obstacle avoidance module to safely reach the current goal selected by PIVOT.

\subsubsection{PACER~\cite{mao2024pacer}} PACER is a terrain-aware model that predicts BEV costmaps from BEV images. It requires pre-enumerating an image context that specifies the preference ordering for terrains. In all environments, we would like the robot to prefer traversing terrains in the following preference order: sidewalk, dirt, grass, rocks. Thus, we condition PACER on this preference order for all environments.
\begin{table}[htbp]
\centering
\caption{Architecture Hyperparameters for \ourmodel{}.}
\label{table:hyp}
\begin{tabular}{l c}
\hline
\textbf{Hyperparameter} & \textbf{Value} \\
\hline
\textbf{RGB-D Encoder} $(f_\text{rgbd})$ &\\
Base Architecture & EfficientNet-B0~\cite{tan2019efficientnet}\\
Number of Input Channels & 4 \\
Input Spatial Resolution & $512 \times 612$ \\
Output Spatial Resolution & $128 \times 153$ \\
Output Channel Dimension & 256 \\
\textbf{Semantic Decoder Head} $(f_\text{semantic})$ &\\
Input Spatial Resolution & $128 \times 153$ \\
Input Channel Dimension & 256 \\
Output Spatial Resolution & $128 \times 153$ \\
Output Channel Dimension & 128 \\
Number of Hidden Layers & 4 \\
\textbf{Depth Completion Head} $(f_\text{depth})$&\\
Input Spatial Resolution & $128 \times 153$ \\
Input Channel Dimension & 256 \\
Output Spatial Resolution & $128 \times 153$ \\
Number of Depth Bins & 128 \\
Number of Hidden Layer & 2 \\
Depth Discretization Method & Uniform \\
Number of Intermediate Layers & 2 \\
\textbf{Lift Splat Module} $(f_\text{splat})$&\\
Total Input Channel Dimension & 288 \\
Input Semantic Channel Dimension & 256 \\
Input Depth Channel Dimension & 32 \\
Map Cell Resolution & $0.1m \times 0.1m \times 3$ \\
Output Channel Dimension & 96 \\
Output Spatial Resolution & $256 \times 256$ \\
\textbf{BEV Inpainting Backbone} $(f_\text{bev})$&\\
Base Architecture & ResNet18~\cite{he2016deep}\\
Input Channel Dimension & 96 \\
Input Spatial Resolution & $256 \times 256$\\
Output Static Panoptic Channel Dimension & 32 \\
Output Dynamic Panoptic Channel Dimension & 32 \\
Output Elevation Channel Dimension & 1 \\
\textbf{Reward Function $(r_\phi)$}&\\
Base Policy Architecture & Value Iteration Network~\cite{tamar2016value}\\
Base Reward Architecture & MultiScale-FCN~\cite{wulfmeier2015maximum}\\
Input Spatial Resolution & $256 \times 256$\\
Input Channel Dimension & 65\\
Prepool Channel Dimensions & [64, 32]\\
Skip Connection Channel Dimensions & [32, 16]\\
Trunk Channel Dimensions & [32, 32]\\
Postpool Channel Dimensions & [48]\\
\# Future Actions & 50 \\
\# Actions Per State & 8\\
Discount Factor & 0.99\\
\hline
\tablabel{modelhyperparameters}
\vspace{-10pt}
\end{tabular}
\end{table}

\begin{table}[htbp]
\centering
\caption{Training Hyperparameters for \ourmodel{}.}
\label{table:hyp}
\begin{tabular}{l c}
\hline
\textbf{Hyperparameter} & \textbf{Value} \\
\hline
\textbf{RGB-D Backbone Training} $(f_\text{rgbd})$ &\\
Semantic Loss Weight $(\alpha_1)$ & 2.0\\
Depth Completion Loss Weight $(\alpha_2)$ & 0.5\\
\# Training Epochs & 50\\
Batch Size & 12\\
Optimizer & AdamW~\cite{loshchilov2017decoupled}\\
Adam $\beta_1$ & 0.9\\
Adam $\beta_2$ & 0.999\\
Learning Rate & $ 5 \times 10^{-3}$\\
Learning Rate Scheduler & Exponential\\
Learning Rate Gamma Decay $\gamma$ & 0.98\\
GPU Training Hours & 10 Hours\\
GPUS Used & 3 Nvidia H100 GPUs\\
\textbf{BEV Inpainting Backbone Training}  $(f_\text{bev})$&\\
Static Panoptic Loss Weight $(\beta_1)$ & 1.0\\
Dynamic Panoptic Loss Weight $(\beta_2)$ & 2.0\\
Elevation Loss Weight $(\beta_3)$ & 3.0\\
Warmup Epochs & 5\\
\# Training Epochs & 50\\
Batch Size & 24\\
Optimizer & AdamW~\cite{loshchilov2017decoupled}\\
Adam $\beta_1$ & 0.9\\
Adam $\beta_2$ & 0.999\\
Learning Rate & $ 5 \times 10^{-4}$\\
Learning Rate Scheduler & Exponential\\
Learning Rate Gamma Decay $\gamma$ & 0.98\\
GPU Training Hours & 24 Hours\\
GPUS Used & 3 Nvidia H100 GPUs\\
\textbf{Reward Function Training} $(r_\phi)$&\\
Batch Size & 30\\
Optimizer & AdamW~\cite{loshchilov2017decoupled}\\
Adam $\beta_1$ & 0.9\\
Adam $\beta_2$ & 0.999\\
Learning Rate & $ 5 \times 10^{-4}$\\
Learning Rate Scheduler & Exponential\\
Learning Rate Gamma Decay $\gamma$ & 0.96\\
Reward Learning Weight $\mathcal{L}_\text{IRL}$& 1.0\\
Reward Smoothness Penalty Weight & 4.0\\
GPU Training Hours & 2 Hours\\
GPUS Used & 3 Nvidia H100 GPUs\\
\hline
\tablabel{traininghyperparameters}
\vspace{-10pt}
\end{tabular}
\end{table}

\subsection{\ourmodel{} Counterfactual Generation Details}
\seclabel{appendix:counterfactualgeneration}
We supplement~\ssecref{rewardfunctionlearning} with additional details regarding the hyperparameters used for generating counterfactual demonstrations. To encourage generating diverse counterfactuals that explore more of the state space, we perturb 3 control points along the expert trajectory according to a non-zero Gaussian distribution. More specifically, for half of the generated trajectories, we sample control points according to a positive mean $\mu=1$ with standard deviation $\sigma=0.5$ to perturb these trajectories to one side of the expert trajectory. We repeat this for the other half, setting $\mu=-1$ and $\sigma=0.5$ to generate trajectories on the opposite side. Additionally, we empirically find that replacing the Hybrid A* path planner by fitting a polynomial line on the control points provides comparable performance gains and is more robust when it is difficult to find a kinematically feasible path between control points.

In total, we generate 10 alternate trajectories for each training sample. On average, we find that providing just 1-5 negative counterfactual annotations for 3\% of the training dataset significantly improves reward learning for $r_\phi$.~\figref{counterfactualannotationtool} presents our counterfactual annotation tool showing qualitative examples of the front view and BEV image observation presented to the human annotator when selecting counterfactual trajectories. We use a fixed set of criteria for annotating counterfactuals, including but not limited to: driving on undesirable terrains, irrecoverable failures like driving off curbs, and unsafe behaviors like colliding into objects and veering off crosswalks. Using our tool, a single human annotator is able to annotate the entire training dataset in approximately 2 hours.

\begin{figure}
    \centering
    \vspace{-10pt}
    \includegraphics[width=\linewidth]{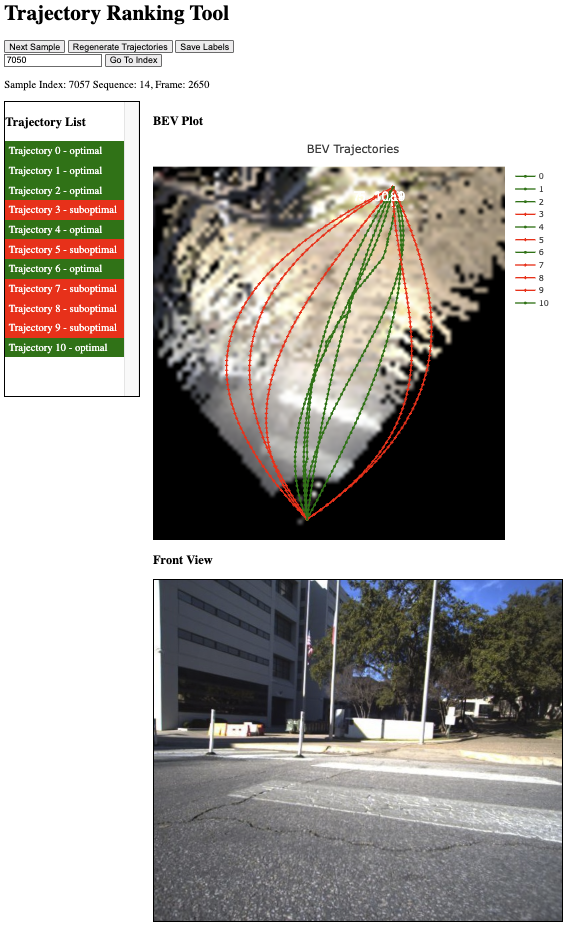}
    \caption{Counterfactual annotation tool used for labelling counterfactuals for training the reward function $r_\phi$. We provide the front view RGB image and BEV RGB image to the human annotator for context. In the image below, the trajectories annotated in red are counterfactuals and the trajectories in green are considered acceptable.}
    \figlabel{counterfactualannotationtool}
    \vspace{-10pt}
\end{figure}

\subsection{Qualitative Short Horizon Experiment Analysis}
\seclabel{appendix:shorthorizonanalysis}

\begin{figure*}[!t]
\begin{center}
    \centering
	\includegraphics[width=\textwidth]{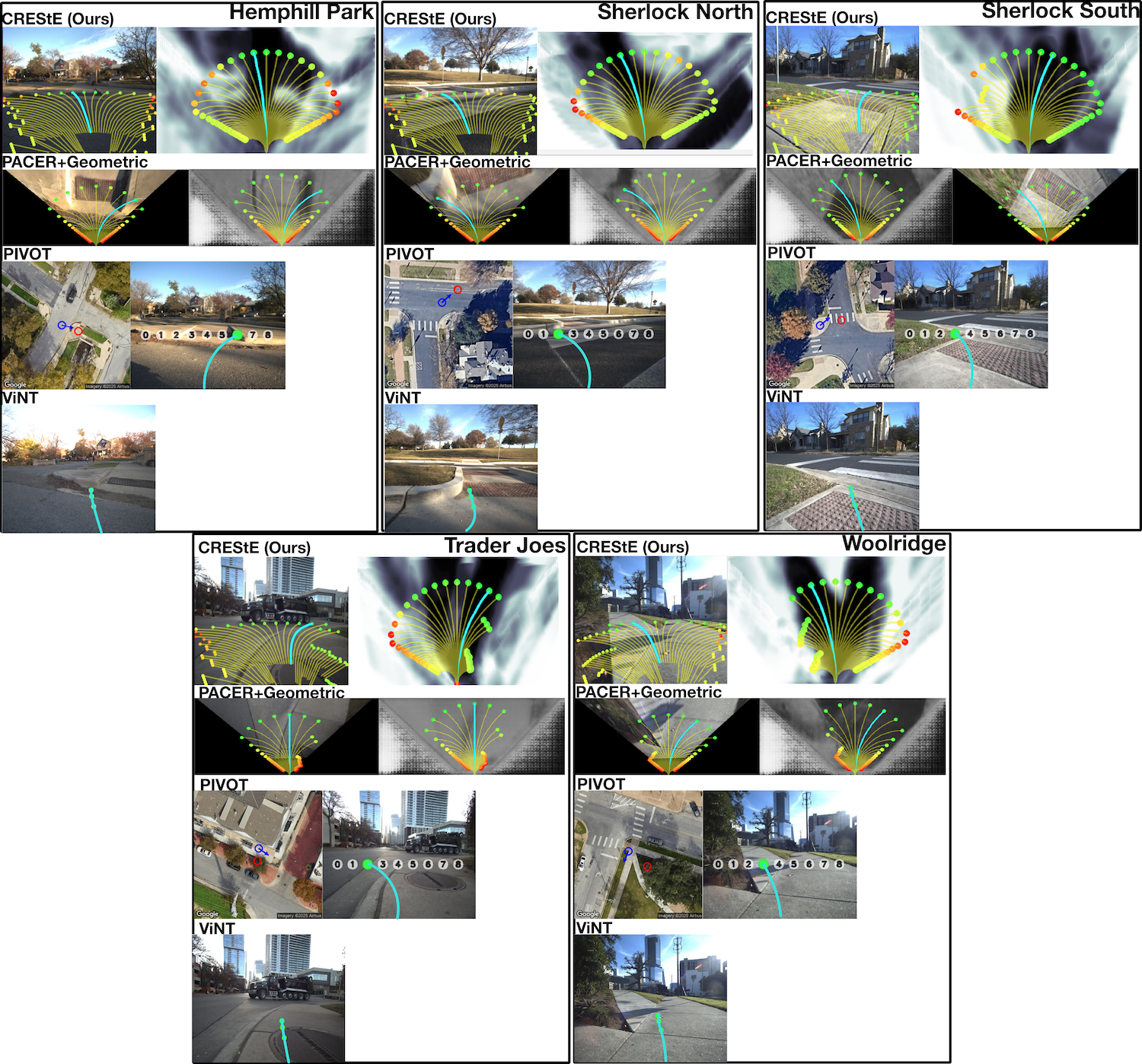}
	\captionof{figure}{Qualitative comparison of local paths planned by different learned baselines in different geoegraphic locations: Hemphill Park, Sherlock North, Sherlock South, Trader joes, Woolridge. We visualize each chosen path either in bird's eye view (BEV) or on the front view RGB image in aqua blue. For PACER+Geometric~\cite{mao2024pacer}, we provide the BEV image used by the model. For PIVOT~\cite{nasiriany2024pivot}, we show the annotated satellite image and front view image used to prompt the VLM. For ViNT~\cite{shah2023vint}, we front view RGB image given as input. For more information regarding each baseline, we refer readers to Appendix~\ssecref{appendix:failureanalysisbaselines}.}
    \figlabel{failureanalysis}
    \vspace{-25pt}
\end{center}%
\end{figure*}
\begin{figure*}[htbp]
\begin{center}
    \centering
	\includegraphics[width=1.0\textwidth]{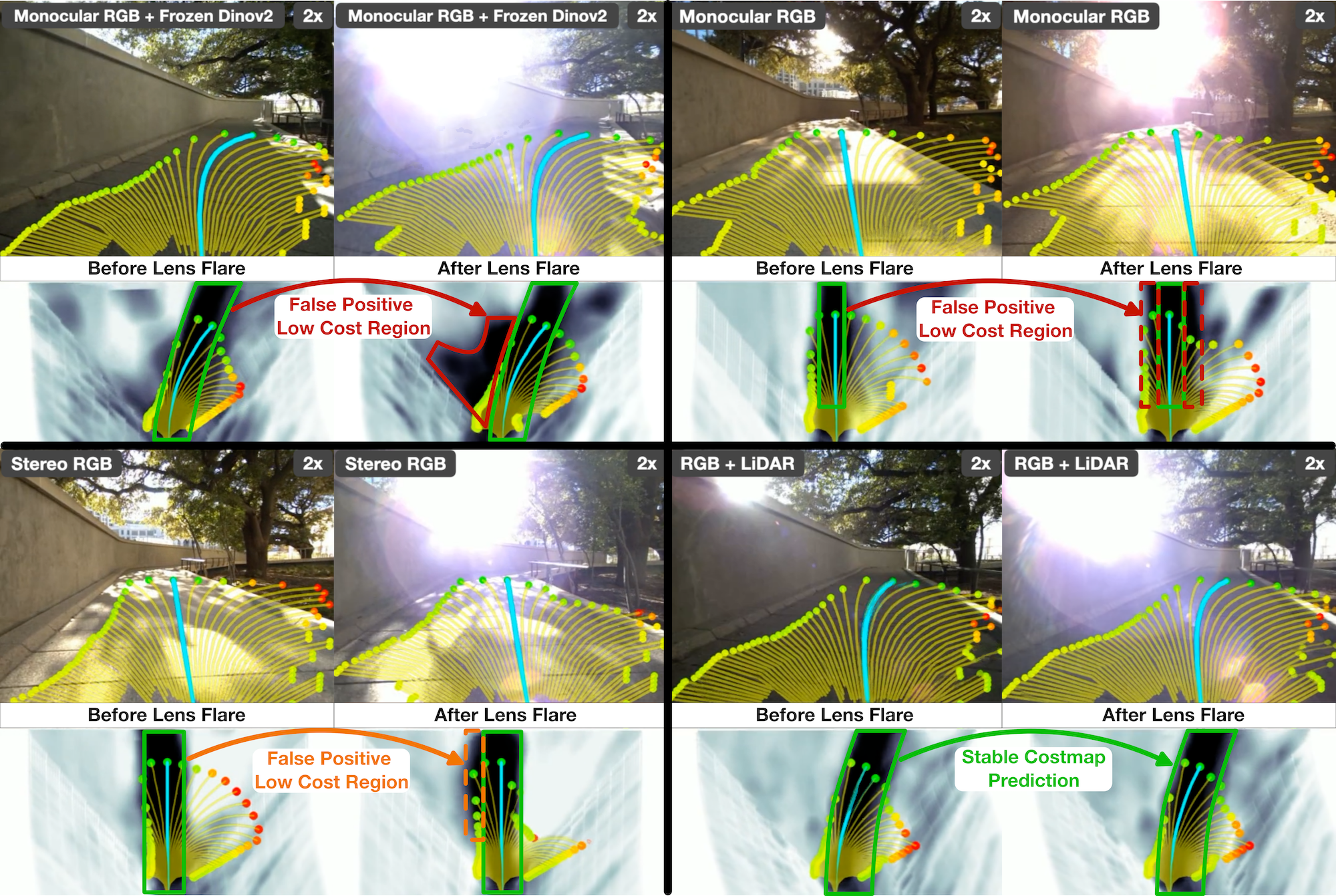}
	\captionof{figure}{Qualitative comparison of different \ourmodel{} architecture variations in geographic location 6, Blanton Museum. These variants consist of: 
    1) \ourmodel{}-RGB-FROZEN (top left), our model but only using monocular RGB inputs and a frozen Dinov2~\cite{oquab2023dinov2} encoder, 2) \ourmodel{}-RGB (top right), our model but only using monocular RGB inputs with a distilled RGB encoder, 3) \ourmodel{}-STEREO (bottom left), our model but only using stereo RGB inputs and an additional stereo RGB backbone~\cite{shamsafar2022mobilestereonet} for depth prediction, 4) \ourmodel{} (bottom right), our original model using monocular RGB and sparse depth from LiDAR as inputs. We visualize the predicted bird's eye view (BEV) costmap and front view RGB image with the chosen trajectory in aqua. For more information regarding each baseline, we refer readers to Appendix~\ssecref{appendix:failureanalysismodality}.}
    \figlabel{failureanalysismodality}
    \vspace{-25pt}
\end{center}%
\end{figure*}

In this section, we present a qualitative analysis of the short horizon experiments for each location. For each experiment area illustrated in~\figref{shorthorizonexperiments}, we compare the planned path for each learned baseline in approximately the same location. We will first compare \ourmodel{} against existing navigation baselines in~\ssecref{appendix:failureanalysisbaselines} before comparing different variations of the \ourmodel{} architecture in~\ssecref{appendix:failureanalysismodality}.

\subsubsection{Comparing \ourmodel{} Against Existing Baselines} \sseclabel{appendix:failureanalysisbaselines}\figref{failureanalysis} provides additional context to understand the inputs given to each baseline and the planned path, which we describe as follows: 1) \ourmodel{} - The input RGB and sparse depth image (not shown) and predicted BEV cost map where darker regions correspond to low cost, 2) PACER+Geometric (PACER+G)~\cite{mao2024pacer} - The input BEV image and predicted BEV cost map where darker regions correspond to lower cost, 3) PIVOT~\cite{nasiriany2024pivot} - The annotated satellite image with the robot's current location (blue circle), heading (blue arrow), and goal location (red circle). The front view RGB image annotated with numbered circles for prompting the VLM along with the chosen circle highlighted in green, 4) ViNT~\cite{shah2023vint} - The front view image annotated with the local path waypoints predicted by ViNT. 

Analyzing each model, we find that PACER+G struggles to infer the cost for terrains that are underrepresented in the training dataset, such as dome mats. Furthermore, when two terrains look visually similar, such as the sidewalk and road pavement for the Trader Joes location, PACER infers the costs incorrectly. PIVOT can select the correct local waypoint in scenes with a large margin for error, but struggles at dealing with curb cuts (Hemphill Park, Sherlock North) and narrow sidewalks (Trader Joes). Long latency between VLM queries negatively impacts the model's ability to compensate for noisy odometry. This effect is particularly apparent in narrow corridors or sidewalks where even minor deviation from the straight line path results in failure. ViNT can successfully maintain course in straight corridors and sidewalks, but struggles to consider factors like curb cuts and terrains. We hypothesize this is because our dataset is not sufficiently large for learning generalizable features from expert demonstrations alone. Furthermore, demonstrations with straight line paths dominate our dataset, making it difficult for behavior cloning methods like ViNT to learn which factors influence the expert to diverge from the straight line path.

\subsubsection{Comparing Different Variations of \ourmodel{}}
\sseclabel{appendix:failureanalysismodality} We evaluate four variations of \ourmodel{} in the same geographic location under adverse lighting conditions in~\figref{failureanalysismodality}. Our monocular RGB only model, \ourmodel{}-RGB, is identical to the original model but does not process sparse depth inputs. In \ourmodel{}-RGB-FROZEN, we replace the original RGB-D encoder with a frozen Dinov2 encoder (VIT-S14), a 21M parameter pre-trained model, and downsample the input images from 512x612 to 210x308. This is essential for performing real-time inference using the onboard GPU. \ourmodel{}-STEREO uses an EfficientNet-B0~\cite{tan2019efficientnet} encoder to extract features from a rectified stereo RGB pair and the MobileStereoNet~\cite{shamsafar2022mobilestereonet} backbone to classify the depth of each pixel. For \ourmodel{}-STEREO, we backproject image features from the RGB backbone $z_\text{rgb}$ from the left image only to maintain a consistent field-of-view with other variations.

We find that \ourmodel{}-RGB and \ourmodel{}-RGB-FROZEN are adversely affected by heavy lighting artifacts. While they are robust to most perceptual aliasing, heavy aliasing (e.g. lens flares) degrades costmap quality. As shown in~\figref{failureanalysismodality}, the monocular RGB only models inflate low cost regions beyond the traversable area under heavy perceptual aliasing. We find that \ourmodel{}-STEREO, our stereo RGB architecture variation, performs more robustly than the monocular RGB only models, but still predicts low cost regions incorrectly at the boundaries between traversable regions. In contrast, \ourmodel{}, which fuses monocular RGB and sparse depth from LiDAR, predicts accurate costmaps even under adverse lighting conditions. We believe this occurs because LiDAR is unaffected by lighting artifacts, allowing the model to compensate for degraded image features.

\end{document}